\documentclass[journal,comsoc]{IEEEtran}

\usepackage[T1]{fontenc}
\usepackage{cite}

\usepackage{subcaption}

\usepackage{tikz}
\usepackage{changes}
\usetikzlibrary{shapes,arrows}
\tikzstyle{block} = [rectangle, draw, fill=blue!20, 
text width=5em, text centered, rounded corners, minimum height=4em, , node distance=3cm]
\tikzstyle{line} = [draw, -latex']

\ifCLASSINFOpdf
  \graphicspath{{./fig/}{./fig_supple/}}
\else
\fi

\usepackage{hyperref}  

\usepackage{amsmath}
\usepackage[cmintegrals]{newtxmath}
\usepackage{url}

\usepackage{bm}
\usepackage{sidecap}
\usepackage{enumitem} 

\usepackage[]{todonotes}

\usepackage{multicol}

\begin{document}

\title{Deep Vehicle Detection in Satellite Video}
\author{Roman~Pflugfelder\textsuperscript{\textsection},
        Axel~Weissenfeld\textsuperscript{\textsection}
        and~Julian~Wagner%
\thanks{R. Pflugfelder is with the Dynamic Vision and Learning Group, Technical University Munich,
e-mail: roman.pflugfelder@tum.de.}
\thanks{A. Weissenfeld  is with the Center
of Digital Safety \& Security, Austrian Institute of Technology,
e-mail: axel.weissenfeld@ait.ac.at.}
\thanks{J. Wagner was with the Institute of Visual Computing \& Human-Centered Technology, University of Technology Vienna,
e-mail: julian.wagner@aon.at.}
}

%



\maketitle

\begingroup\renewcommand\thefootnote{\textsection}
\footnotetext{Both authors contributed equally to this article.}
\endgroup

\begin{abstract}
This work presents a deep learning approach for vehicle detection in satellite video. Vehicle detection is perhaps impossible in single EO satellite images due to the tininess of vehicles (4-10 pixel) and their similarity to the background. Instead, we consider satellite video which overcomes the lack of spatial information by temporal consistency of vehicle movement. A new spatiotemporal model of a compact $3 \times 3$ convolutional, neural network is proposed which neglects pooling layers and uses leaky ReLUs. Then we use a reformulation of the output heatmap including Non-Maximum-Suppression (NMS) for the final segmentation. Empirical results on two new annotated satellite videos reconfirm the applicability of this approach for vehicle detection. They more importantly indicate that pre-training on WAMI data and then fine-tuning on few annotated video frames for a new video is sufficient. In our experiment only five annotated images yield a $F_1$ score of 0.81 on a new video showing more complex traffic patterns than the Las Vegas video. Our best result on Las Vegas is a $F_1$ score of 0.87 which makes the proposed approach a leading method for this benchmark.
\end{abstract}


\ifCLASSOPTIONpeerreview
\begin{center} \bfseries EDICS Category: 3-BBND \end{center}
\fi
%
\IEEEpeerreviewmaketitle

\section{Introduction}
\IEEEPARstart{O}{Object} detection in visual data is a very important and still unsolved problem. For example, the problem becomes challenging in aerial imaging and remote sensing as the scenes and the data differ significantly from the case usually considered in computer vision~\cite{benenson-cvpr2013, ren-nips2015}.
Remote detection is important in surveillance, as demanding applications let surveillance currently undergo a transition from short ranges (as with classical security cameras) to sceneries such as whole cities and traffic networks but also to rural environments such as forests. New, low orbit satellite constellations will allow multiple daily revisits and constantly falling costs per image. Innovations are possible in intelligent transportation systems, e.g. by large-scale traffic monitoring for better urban planning. Military and civilian security is another area to benefit with applications including military reconnaissance, border surveillance and search \& rescue.

\begin{figure}
	\centering
	\includegraphics[width=0.45\textwidth]{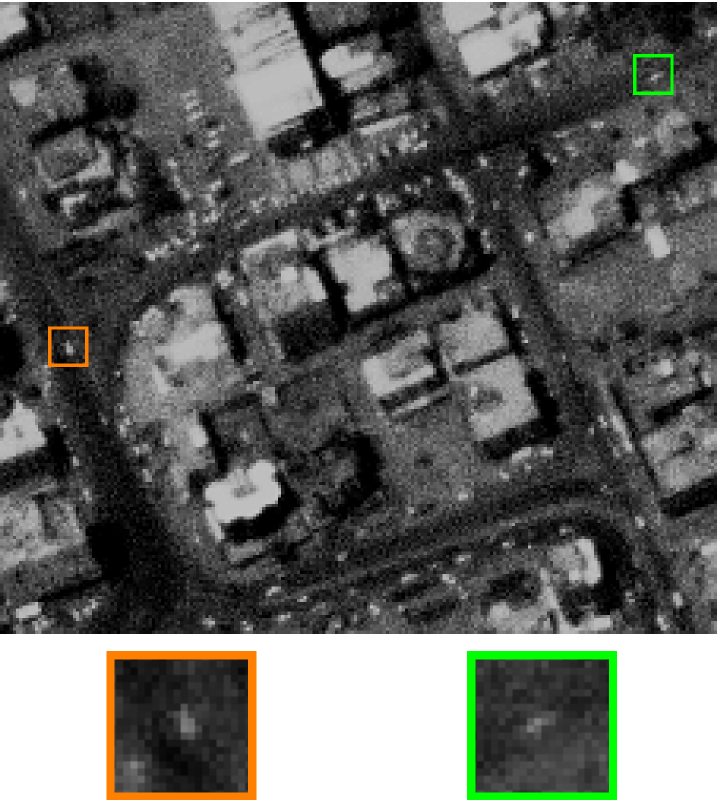}
	\caption[carCap]{Object detection is challenging due to the tiny objects (four to ten pixels) and the large number of objects in an image. Top: Small region cropped and enlarged from a satellite video frame. Bottom: Two enlarged patches of two vehicles (orange, green) in the region.}
	\label{fig:intro:challenge} 
\end{figure}

\begin{figure}
\centering
\includegraphics[width=0.48\textwidth]{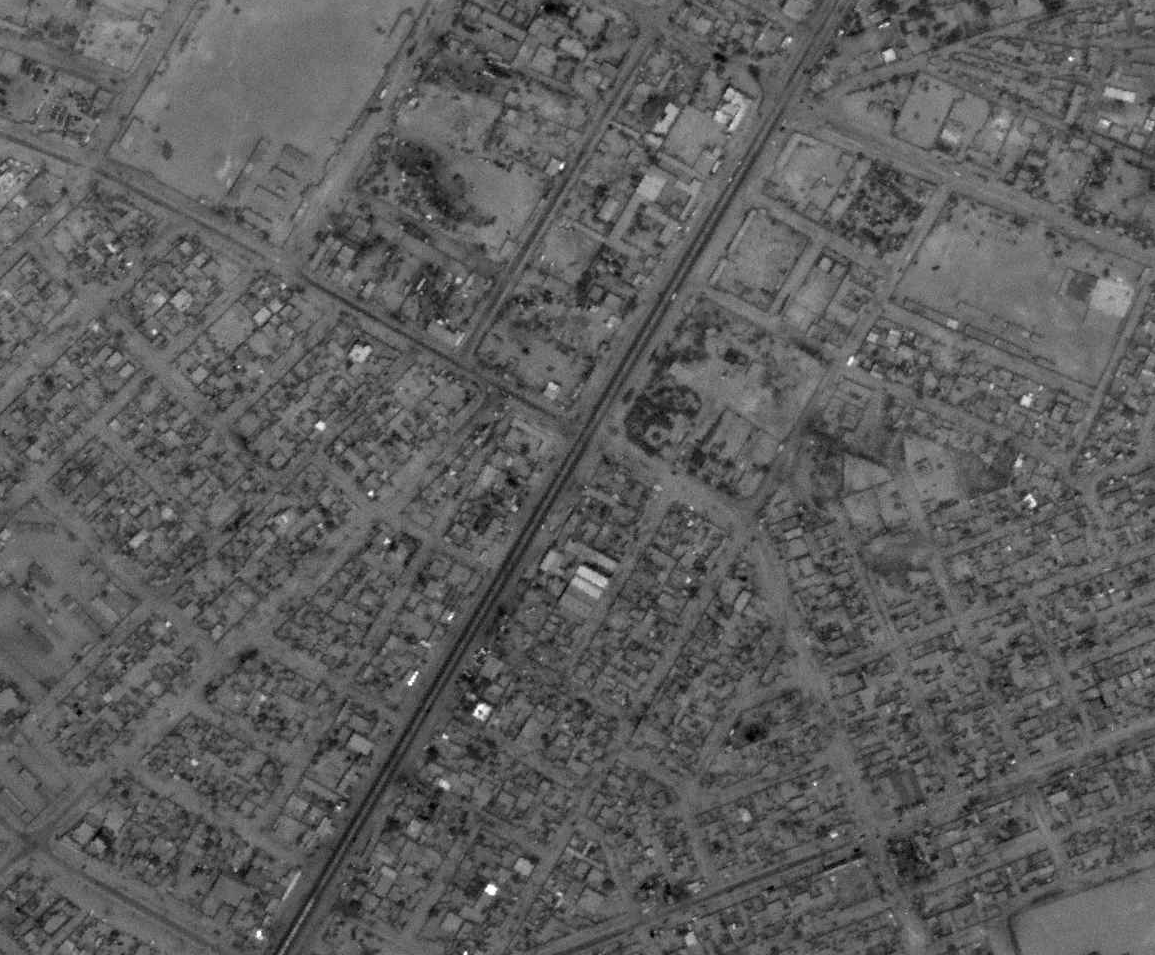}
\caption{Object detection is challenging due to the extent of the scenery captured by a single satellite image. As example a video frame of Planet's SkySat video product, which depicts the center of Agadez (2\,km\textsuperscript{2}). Agadez with a population of around 110,000 is in Niger.}
\label{fig:exampleAgadez}
\end{figure}

Although the overhead view of remotely acquired data reduces occlusion and perspective distortion, new difficulties arise when compared with classical short-range surveillance. Typical satellite images are very large in resolution and data size. For example, Planet's SkySats capture an area up to $5.9\times20$\,km\textsuperscript{2} with 50\,cm to one meter ground sample distance (GSD) depending on their relative position to nadir and the computational pre-processing. By that objects such as vehicles reduce in pixel size {\em by orders of magnitude} from 10\textsuperscript{4}\,pixel as for a security camera to 4-10\,pixel~\cite{pflugfelder2020learning} in satellite images depending on the camera's GSD.
This severe magnification of scenery {\em and} reduction of object size to very tiny appearances of a few pixel has consequences to visual recognition. Object detection becomes very ambiguous and sensitive to noise and the search space dramatically increases and becomes very sparse (see Fig.~\ref{fig:intro:challenge} and Fig.~\ref{fig:exampleAgadez}). Inferred labels of data usually capture sole positions instead of bounding boxes or contours, as the extent of objects is even for humans unrecognizable. Furthermore, a satellite image potentially captures thousands of objects compared to a few targets in classical surveillance. All this leads to major difficulties if not inapplicability of vanilla methods~\cite{lalonde-cvpr2018} in computer vision.

This work aims at the small size of vehicles. The reader is referred to LaLonde et al.~\cite{lalonde-cvpr2018} for a rigorous discussion on the challenge of large image sizes. Instead of high-resolution satellite images~\cite{imbert-thesis2019} the idea is to use satellite video for vehicle detection. Video offers compared to single images high spatiotemporal resolution and information. 


We extend previous work\cite{pflugfelder2020learning} which showed with 0.84 $F_1$ score for the AOI\,1 (area of interest) on the Las Vegas video very promising results when compared to other state-of-the-art methods e.g. E-LSD~\cite{zhang-corr2019b}. The assumptions of this approach are vehicle movement and smooth trajectories which enable the deep learning to exploit these inductive biases by a spatiotemporal, neural network (FoveaNet) which was proposed by LaLonde et al.~\cite{lalonde-cvpr2018} for Wide-Area Motion Imagery (WAMI). WAMI is usually acquired by airborne light-field cameras. For example, WPAFB 2009~\cite{WPAFB} provides stitched 25\,cm GSD, 16\,bit panchromatic, $26k \times 21k$\,pixel images at 1.25\,Hz.
The approach~\cite{pflugfelder2020learning} shows challenges with small distances between neighboring vehicles (usually a few pixels). In this case FoveaNet~\cite{lalonde-cvpr2018} creates a heatmap with a large number of connected regions, which finally results in a large number of false negative detections.

We propose in this work an improved Foveanet architecture and a new approach to the post-processing which significantly increases the accuracy. Finally, the work considers the transfer learning on new video with a reduced number of hand-labelled samples. In more detail, this work contributes to the literature,
\begin{enumerate}
    \item by accumulating the target Gaussian heatmaps for training with the maximum for each location instead of taking the sum, and by replacing in the segmentation Otsu thresholding~\cite{lalonde-cvpr2018} with NMS~\cite{bodla2017soft} which both tempers the shortcomings of our initial detector~\cite{pflugfelder2020learning} in case of proximate vehicles,
    
    \item by additionally proposing FoveaNet4Sat with improvements to the FoveaNet\cite{lalonde-cvpr2018} architecture such as $3\times 3$ convolutions~\cite{pflugfelder2020learning}, neglected max-pooling and leaky ReLUs~\cite{pflugfelder2020learning} which e.g. on Las Vegas AOI 1 yield a gain in performance from 0.75~\cite{wagner-phd2020} to 0.88 in $F_1$ score (+17.3\,\%),
    
    \item by a new video with 87,274 annotated vehicles in the city of Khartoum, Sudan and by empirical results on this new video that reconfirm the improvement of the proposed method when compared with LaLonde et al.~\cite{lalonde-cvpr2018}\footnote{Upon acceptance by the owners, the Khartoum video and annotation will be made available with this article under an open source license.}, and
    
    \item by new empirical results on the number of training samples needed to surpass 0.8 $F_1$ score on Khartoum. The proposed method needs in the best case only five hand-labelled images (five heatmaps for $5 \times 5$ training images), far less than LaLonde's method which needs 238 samples (-97.9\,\%).
\end{enumerate}




The remainder of this article is organised as follows: Section~\ref{sec:rel-work} summarises literature on vehicle detection and tracking for satellite video. We also compare this literature to other work on single satellite images and aerial images. The proposed method is detailed in Section~\ref{sec:method}. The new Khartoum and Agadez videos as well as the experimental setup including Las Vegas are summarised by Section~\ref{sec:exp-setup}. Section~\ref{sec:results} then gives a detailed analysis on FoveaNet4Sat including the empirical results on the number of training samples. Final conclusions and suggestions for future work are given in Section~\ref{sec:conclusion}.

\begin{figure*}[t]
\begin{center}
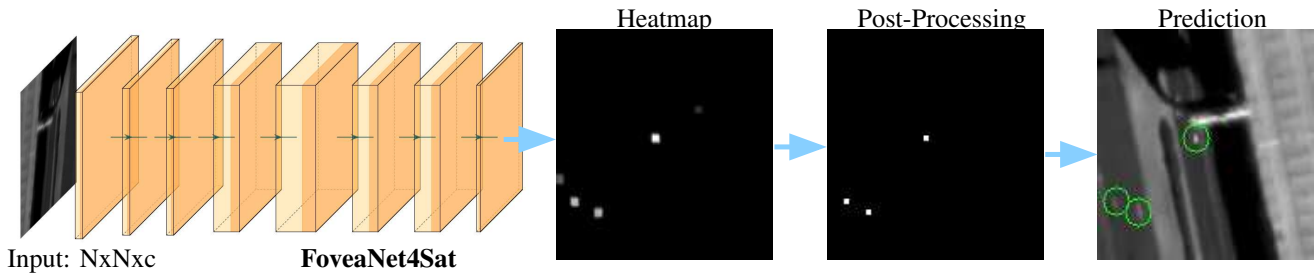 
\caption{The object detection process has two steps: $c$ $N \times N$ images stacked together are the input to FoveaNet4Sat. The network predicts a heatmap which indicates the likelihood that an object is at a given image coordinate. A post-processing step derives the object locations (green circles) based on NMS.}
\label{figure:overview_foveanet4sat}
\end{center}
\end{figure*}

\section{Related Work}
\label{sec:rel-work}
Recent literature~\cite{mou-igarss2016, mundhenk-eccv2016, zhang-spie2017, koga-rs2018, yang-corr2018, ding-jprs2018, guo-rs2018, lalonde-cvpr2018, zhang-rsens2019, imbert-thesis2019, AL-SHAKARJI_2019_CVPR_Workshops, yang-iccv2019} suggest to apply deep learning for vehicle detection on aerial images. E.g. Mundhenk et al.~\cite{mundhenk-eccv2016} show with $F_1$ scores larger than 0.9 excellent results on a reasonable dataset. Their idea is to pixel-wise classify vehicle vs. background by combining Inception~\cite{Szegedy_2015_CVPR} and ResNet~\cite{he-cvpr2016} to give a heatmap. They assume a fixed vehicle size and use NSM for semantic segmentation. They, however, report moderate performances for GSDs larger than 15\,cm.
Imbert~\cite{imbert-thesis2019} proposes a generative U-Net in combination with hard negative mining for single, high-resolution satellite images (Worldview). Unfortunately, quantitative results are kept confidential. It is therefore unclear, if vehicle detection in single images is possible, given the limitations of the data.
Another issue of the still image is the impossibility to capture the dynamic behaviour of vehicles which is essential for many applications. For example, vehicle heading and speed are important indicators in traffic models. Although rapid re-targeting for multi-angular image sequences with Worldview-2 is possible~\cite{meng}, the time interval of around one minute between consecutive images is too large for reasonable analysis.

Satellite video was introduced 1999 by DLR-TubSat, since 2013 Planet's SkySat-1 delivers up to 120\,s, 50\,Hz, 2K panchromatic video. China's Jilin programme was launched in 2015 and now provides 4\,MP RGB 30\,Hz video that is commercially offered by Chang Guang Satellite Technology. Research satellites like Lapan-TubSat (2007), Tiantuo-2 (2014) are operating but data is not available to the public. Urthecast operated since 2014 the 8\,MP 30\,Hz, 1\,m GSD RGB Iris camera at ISS space station, but finished its service before 2020. Earth-i operated a prototype satellite Carbonite-2 (2018) and plans its Vivid-i programme for 2022.

The labelling of video is very tedious, therefore research on vehicle detection in satellite video mostly relies on background subtraction (BGS)~\cite{zhang-spie2017, kopsiaftis-igarss2015, xu-jrsgis2017, yang-sensors2016, ahmadi-ijrs2019, chen-access2020} and frame differencing (FD)~\cite{li-JARS2019, ao-corr2018, ao-tip2019}, except Al-Shakarji et al.~\cite{AL-SHAKARJI_2019_CVPR_Workshops} who combined YOLO~\cite{redmon-corr2018} with spatiotemporal filtering on WAMI ($F_1$ score of 0.7), and Mou and Zhu~\cite{mou-igarss2016} who use KLT tracking on video with a SegNet~\cite{Badrinarayanan-tpami2017} on overlapping multispectral data, however, they did not show results for vehicles. Zhang and Xiang~\cite{zhang-spie2017} apply a ResNet~\cite{he-cvpr2016} classifier trained on CIFAR~\cite{krizhevsky-2009} on proposals from a mixture of Gaussians (foreground model), but did not show a proper evaluation.
Given the temporal changes, the detectors apply either connected component analysis~\cite{kopsiaftis-igarss2015, xu-jrsgis2017}, or saliency analysis, segmentation~\cite{yang-sensors2016, li-JARS2019}, or distribution fitting~\cite{ao-corr2018, ao-tip2019}, finally followed by morphology. $F_1$ scores of larger than 0.9 for ships and scores between 0.6 and 0.8 for vehicles on the Burji Khalifa~\cite{yang-sensors2016}, Valencia~\cite{ao-corr2018, ao-tip2019} and Las Vegas~\cite{kopsiaftis-igarss2015} videos suggest BGS, FD for larger objects. Both BGS and FD depend heavily on registration and parallax correction, hence, these methods introduce various nuisances which are difficult to handle.
Recently, BGS is combined with adaptive pre-filtering and with a shallow, convolutional, fully-connected network to regress within $32\times 32$ patches the probability of vehicle appearance~\cite{chen-access2020, chen-spie2021}. The authors report with a $F_1$ score of 0.965 excellent results on the Las Vegas video. However, the results are not directly comparable to the results of Zhang et al.~\cite{zhang-corr2019a, zhang-tgars2020}, Pflugfelder et al.~\cite{pflugfelder2020learning} and this work, as they restricted the evaluation of their method to ten frames of a larger region (Data\,1~\cite{chen-access2020}) including AOI\,1. Furthermore, their evaluation is based on their own labelling.

A different line of recent work~\cite{zhang-corr2019a, zhang-tgars2020, zhang-tpami2021} suggests a subspace approach for discriminating vehicles and background by using low-rank and structured sparse decomposition. The idea shows potential with $F_1$ score results of larger than 0.8 on the Las Vegas video, however it is an open question how their approach works on video without preprocessing~\cite{murthy2014skysat} and with more variability in the traffic pattern as shown in our Khartoum video. An online variant~\cite{zhang-tgars2020} of the method and a recent consideration of the satellite's ego-motion~\cite{zhang-tpami2021} have been proposed.

Most comparable to our work is the recently proposed key-point based detection (CKDNet)~\cite{feng-isprs2021} using an Hourglass backbone and recurrent network for spatiotemporal analysis. Two regressed corners define a bounding box in their work. Although results with a $F_1$ score of 0.88 are comparable to our results (0.88, Tab.~\ref{tab:exp6_1}), this method is by far more computational demanding than FoveaNet4Sat which solves a simpler problem by regressing sole position instead of a bounding box. Furthermore, a direct comparison of the quantitative results is difficult as Zhang et al.~\cite{zhang-corr2019a, zhang-tgars2020, zhang-tpami2021} and we use the same positional instead of Feng et al.'s bounding box labels for annotation.

\section{Method} 
\label{sec:method}
We propose FoveaNet4Sat (FoveaNet for satellite), a convolutional neural network (CNN) for vehicle detection in satellite video (Fig.~\ref{figure:overview_foveanet4sat}). It is inspired by Lalonde et al.~\cite{lalonde-cvpr2018}, who proposed FoveaNet for WAMI.

{\bf 2D vs. 3D CNNs:} FoveaNet and FoveaNet4Sat are both 2D CNNs, i.e. the convolutional kernels are defined in the image domain. There is work in visual action recognition~\cite{Baccouche_2011_HBU, Luo_2019_ICCV, Long_2019_CVPR} that suggest to use 3D networks. 3D CNNs are a generalisation of 2D networks and define a three-dimensional kernel which operates in the image domain and in the time domain of a video. Why are LaLonde et al. and we then able to claim that FoveaNet and FoveaNet4Sat are spatiotemporal networks? The reason lies in the implementation of the 2D CNN and that we use the channels to express the individual images of the video sequence. The convolution is e.g. in PyTorch~\cite{paszke-neurips2019} implemented in a way so that a 2D convolution has independent kernel weights across the image domain and the channel domain. This is the special case of a 3D convolutional kernel and network where the kernel size in the temporal domain is equal to the length of the image sequence which is the input to the network. This approach is limited to grayscale images but is applicable in our case as satellite video is panchromatic.

Karpathy et al.~\cite{karpathy2014large} discussed for this case also models of late fusion where in the first layers of the network the kernels are grouped to individual frames~\cite{paszke-neurips2019}. This idea has been shown earlier for human tracking~\cite{Fan_2010_TNN} where the later spatiotemporal binding is beneficial. The reason is that temporal relationships between spatial features with larger receptive fields make sense for larger objects. In our case, vehicles are tiny therefore early fusion in the first convolutional layer is useful to capture small spatiotemporal patterns.

{\bf Can we interchangeably use for our task 2D networks with early, late binding and 3D networks?} We observe contradicting arguments in literature~\cite{tran2015learning, lalonde-cvpr2018, karpathy2014large} and it is unclear if 2D networks on grayscale video are equally powerful than 3D networks. We leave this to future work. For our case, the number of images is with three to seven rather low. For such a case, the size of the 3D kernel in the temporal domain will be the same as the number of frames. But this leads us exactly to a 2D network with a channel group of one which is the model of FoveaNet and FoveaNet4Sat.

{\bf Registration:} Both WAMI and satellite videos show not only the motion of individually moving targets but also the ego-motion of the sensor platform, i.e. the motion of the satellite or, e.g airplane in case of WAMI.
In order to learn motion information of moving, individual targets both methods (Foveanet and Foveanet4Sat) require that consecutive frames are registered to each other and thereby removing ego-motion of the imaging platform.

\subsection{FoveaNet}
The FoveaNet~\cite{lalonde-cvpr2018} consists of eight convolutional layers. The input to the network is not a single frame but a stack of $c$ frames, where $c$ depicts the number of consecutive adjoining frames in a stack. Hereinafter we refer to $c$ as channels. Thereby, the CNN learns to predict the positions of objects in the mid frame of the overall sequence. All convolutional layers have ReLU activation functions. The number of filters per convolution are 32, 32, 32, 256, 512, 256, 256 and 1 with a kernel size of 15-13-11-9-7-5-3-1. After the first convolution a $\medmuskip=0mu 2 \times 2$ max pooling is carried out. During training the 6$^{th}$ and 7$^{th}$ convolutional layers have a 50\% dropout. The output of the network is a heatmap, which is generated by the final $\medmuskip=0mu 1 \times 1$ convolutional layer where each neuron gives a vote of the likelihood of a moving vehicle at pixel level. Afterwards,  the objects' positions need to be determined from the predicted heatmap. For this, the heatmaps are converted into segmentation maps via Otsu thresholding~\cite{otsu-tsmc1979}. If the segmented area is larger than a threshold, then the center of the area is defined as an object position.


Ground-truth heatmaps are created by placing 2D Gaussians at the locations of the centers of vehicles. Formally, these heatmaps are generated by the following formula:

\begin{equation}
H_{\text sum}(x,y) = \text{min}\left( \sum_{n=1}^{V} \frac{1}{2\pi \sigma^{2}}e^{-\frac{\left( \frac{x-x_{n}}{2^{d}}\right)^2 + \left(\frac{y-y_{n}}{2^{d}}\right)^{2}}{ 2\sigma^{2} }}, 1\right)
\label{heatmap_sum}
\end{equation}

where $V$ is the number of vehicles in the given mid frame, $d$ is the factor of downsampling the heatmap corresponding the the CNN output size and $\sigma$ is the variance of the Gaussian blur fit to the approximate size of target objects in the dataset. $(x, y)$ represent the current position in $H_{\text sum}$ while $(x_{n}, y_{n})$ is the center position of the $n^{th}$ vehicle. The values in $H_{\text sum}$ are clipped at one so clusters of multiple objects are not over-weighted.

\subsection{FoveaNet4Sat}


The architecture of FoveaNet4Sat is based on the FoveaNet but there are some significant differences to adapt the network to the detection of vehicles in satellite video. FoveaNet4Sat has also eight convolutional layers and the input to the network are image stacks. The number of filters per convolution remain the same but we reduce the kernel sizes to 3-3-3-3-3-3-3-1 in analogy to the VGGNet~\cite{simonyan2014very}. The receptive field covered by the deep composition of smaller kernels at each layer is still sufficient to cover the offset of a moving vehicle in two consecutive frames which is necessary to learn valid, spatiotemporal features~\cite{wagner-phd2020}. On the other hand, the reduction of the filter sizes significantly lowers the trainable parameters of the network from $\sim$11.3 million to $\sim$3.4 million, which in turn reduces training time and likely also the number of training data needed. The original FoveaNet has ReLU activation functions. We discovered, however, the problem known as the "Dying ReLU" problem~\cite{he-iccv2015}. Hence, we replace the ReLUs with Leaky ReLUs~\cite{pflugfelder2020learning}, that provide fast convergence during training. By eliminating the pooling layer, heatmaps with full image resolution are generated without loosing accuracy~\cite{springenberg-iclrw2015}. One  advantage of pooling, namely the reduction of dimensions of the feature maps, does not come into play, since the Foveanet4Sat is rather small. Furthermore, max pooling can increase the invariance to
spatial perturbations. In our case, however, the improvement of the invariance does not seem to be as important as a higher resolution of the heatmap, which enables a better detection of vehicles, when their distance from each other is small.




Instead of accumulating the Gaussian distributions as for $H_{\text sum}(x,y)$, we propose to take pixel-wise the maximum of the density values. Consequently, the value at pixel location $(x,y)$ is defined as,

\begin{equation}
H_{\text max}(x, y) = \max_{\forall n \in \{1,...,V\}} \frac{1}{2\pi \sigma^{2}}e^{-\frac{ (x-x_{n})^2 + (y-y_{n})^{2}}{ 2\sigma^{2} }}.
\label{eq:heatmap_max}
\end{equation}

In this way, the values of the heatmap always remain in the range of $(0, \frac{1}{2\pi \sigma^{2}}]$ independent of the total number of objects $V$ in the scene. This prevents clusters of multiple objects from being
over-weighted similar to the clipping to one for $H_{\text sum}$. However the clipping of $H_{\text sum}$ can lead to cut-off maxima, so that the precise information of the vehicles' location is no longer available which is not the case for $H_{\text max}$.





\begin{figure}[!htb]
    \centering
    \subcaptionbox{Pred. heatmap.\label{subfig-1:platzhalter}}{%
      \includegraphics[width=2.8cm]{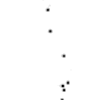}
    }%
    \subcaptionbox{Otsu thresholding.\label{subfig-2:platzhalter}}{%
      \includegraphics[width=2.8cm]{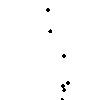}
    }%
    \subcaptionbox{Detections.\label{subfig-3:platzhalter}}{%
      \includegraphics[width=2.8cm]{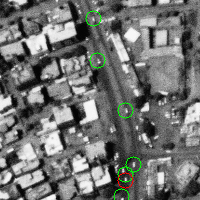}
    }%

    \bigskip  
    \subcaptionbox{Pred. heatmap.\label{subfig-4:platzhalter}}{%
      \includegraphics[width=2.8cm]{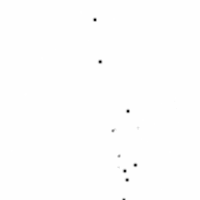}
    }%
    \subcaptionbox{NMS filtering.\label{subfig-5:platzhalter}}{%
      \includegraphics[width=2.8cm]{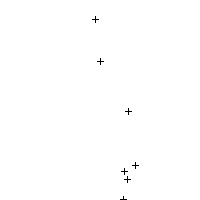}
    }%
    \subcaptionbox{Detections.\label{subfig-6:platzhalter}}{%
      \includegraphics[width=2.8cm]{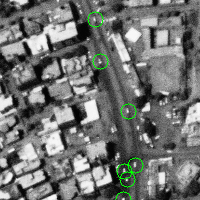}
    }%

    \caption{Comparison of the post-processing based on Otsu thresholding (Fig.~\subref{subfig-1:platzhalter},~\subref{subfig-2:platzhalter},~\subref{subfig-3:platzhalter}) and NMS filtering (Fig.~\subref{subfig-4:platzhalter},~\subref{subfig-5:platzhalter},~\subref{subfig-6:platzhalter}) in heavy traffic. Dense traffic may result in connected regions of the predicted heatmap (e.g. Fig.~\subref{subfig-1:platzhalter}) using Otsu thresholding (Fig.~\subref{subfig-2:platzhalter}). This in turn causes false negative detections (Fig.~\subref{subfig-3:platzhalter}). On the other hand, NMS filtering is able to separate connected regions (Fig.~\subref{subfig-5:platzhalter}) and the model achieves better detection results (Fig.~\subref{subfig-6:platzhalter}). Note, for better visualization we have marked the point detections with crosses in Fig.~\subref{subfig-5:platzhalter}.  (true positive: green, false positive: blue, false negative: red)}
    \label{fig:NMSversusOtsu}
  \end{figure}

The post-processing based on NMS~\cite{bodla2017soft} has two steps: First, a $3 \times 3$ max-pooling kernel is applied to the heatmap followed second by a filtering with a given threshold. The resulting pixels with non-zero values indicate positions of detected vehicles. The advantage of using $H_{\text max}$ and NMS versus $H_{\text sum}$ and Otsu thresholding is particularly noticeable in dense traffic as depicted in Fig.~\ref{fig:NMSversusOtsu}.


\subsection{Transfer learning}
Despite the recent emergence of a new satellite video dataset~\cite{zhao-tgars2022}, to the best of our knowledge there is currently no dataset of sufficient size for deep learning publicly available. In contrast, there are some labelled WAMI datasets accessible; e.g. the WPAFB 2009 dataset~\cite{WPAFB} contains over 2.4 million annotated vehicles. WAMI differs considerably from satellite video. For instance, the WPAFB images have about four times higher GSD than the Las Vegas video.

Our core idea is to use transfer learning for a domain transfer from WAMI to satellite video. For this, we pre-train our CNN on the WPAFB data with a lower image resolution~\cite{pflugfelder2020learning}. Afterwards we fine-tune the CNN on the satellite video. By this, the temporal resolution of the training samples is adopted to the frame-rate of the WPAFB data by skipping frames, i.e. every 10th consecutive video frame is stacked in a training sample.

\begin{figure}[b]
\centering
\includegraphics[width=8.5cm]{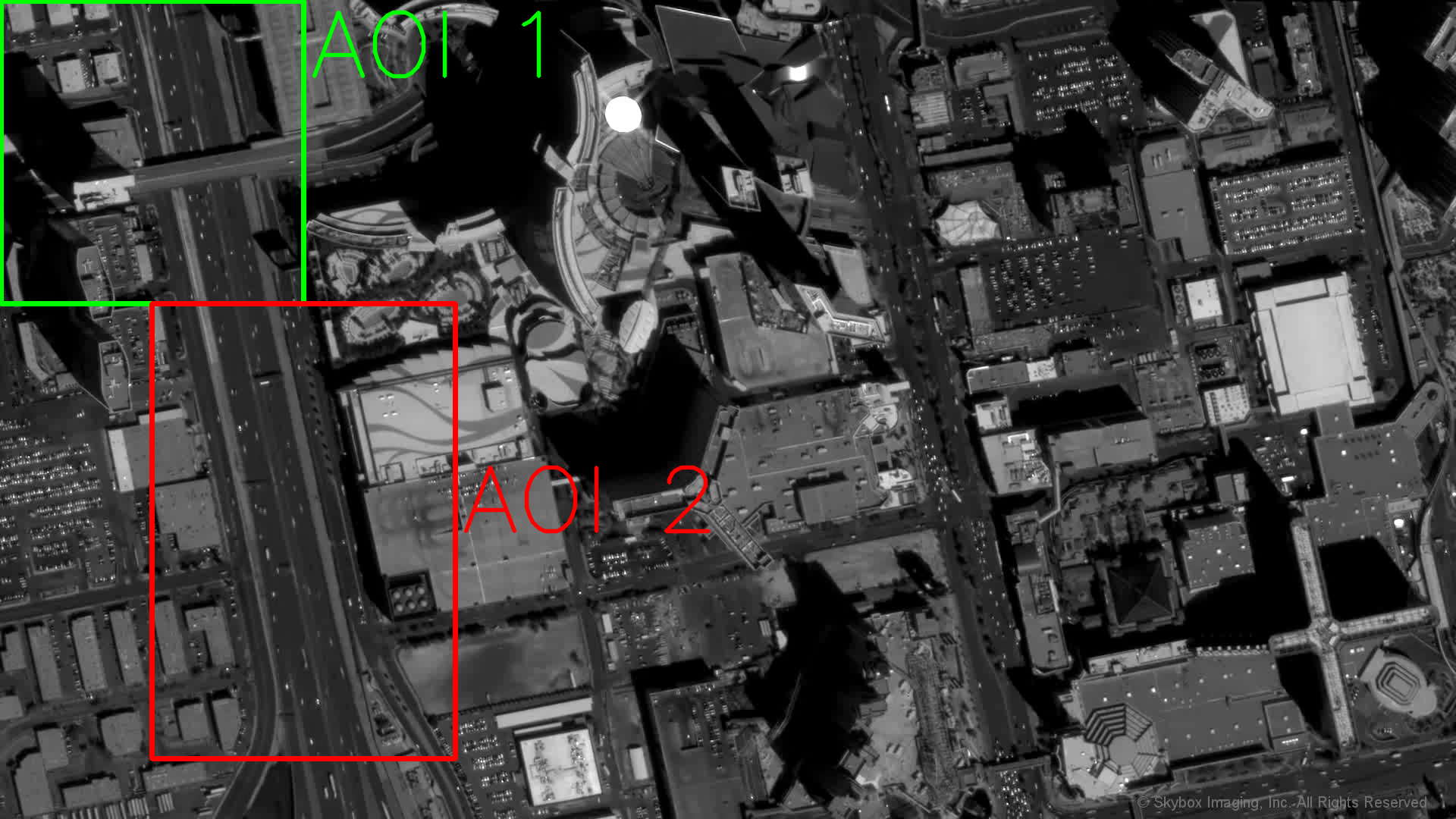}
\caption{First video frame of the Las Vegas video in which AOI\,1 ($400\times400$\,pixel) and AOI\,2 ($600\times400$\,pixel) are shown.}
\label{fig:lasvegas}
\end{figure}

\section{Experimental Setup}
\label{sec:exp-setup}



\begin{figure*}[t]
\centering
\includegraphics[width=1.0\textwidth]{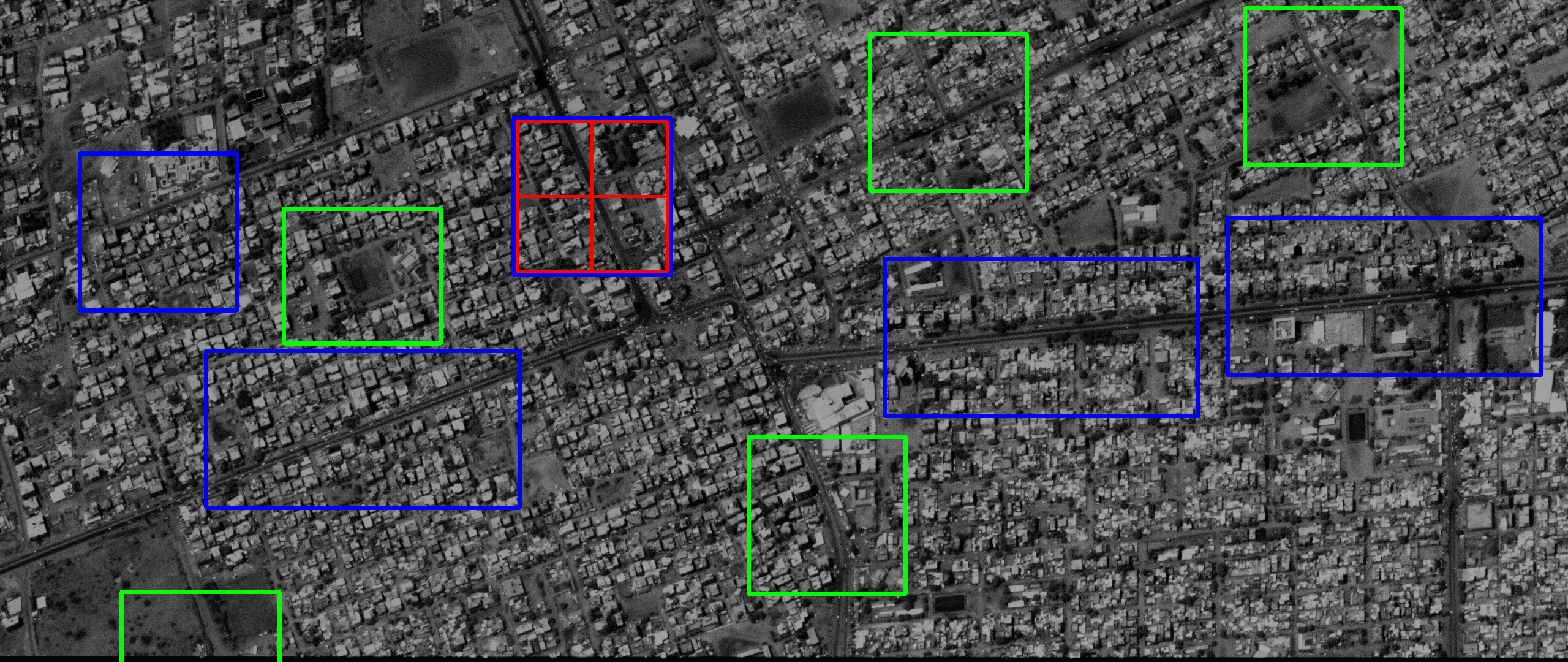}
\caption{First video frame of the Khartoum video. Khartoum is the capital of Sudan with a population of over 5 million inhabitants. AOIs used for training (blue) and evaluation (green) are shown as rectangles. An AOI is split in ROOBIs for the training. Four ROOBIs of a specific AOI are shown in red.}
\label{fig:khartoum}
\end{figure*}

Experiments were performed on four different datasets: the WPAFB 2009 dataset~\cite{WPAFB} and the public SkySat-1 Las Vegas video\,\footnote{\url{https://www.youtube.com/watch?v=lKNAY5ELUZY}} as well as two new satellite videos from Khartoum and Agadez acquired and annotated by us. An overview of the four datasets is provided in Tab.~\ref{tab3}. 

\subsection{Datasets}
The 1025 annotated frames in the WPAFB 2009 dataset depict a part of the Wright Patterson Air Force Base and the surrounding neighborhood. Individual frames are captured with a single large-format, monochromatic, electro-optical sensor comprised of a matrix of six cameras \cite{Cohenour2015}. The GSD of the WPAFB 2009 dataset is about 0.25\,m. The average vehicle size is $18 \times 9$\,pixel according to \cite{lalonde-cvpr2018}. In order to match the GSD of satellite video, which is about 1\,m, a downscaling factor of 0.2 is applied. This downscaling factor does not exactly result in a GSD of 1\,m but it is found that it matches the average vehicle size of $3.6\times1.8$\,pixel as reported by Zhang et al. \cite{zhang-GSRS2019}. In order to remove ego-motion, consecutive frames are registered to each other using the approach described in Sec.~\ref{sec:registration}. For pre-training we selected three AOIs of the WPAFB dataset - 34, 40 and 41 as described in more detail in~\cite{pflugfelder2020learning}. Only frames with moving vehicles are included for training. A vehicle is defined as moving if it moves at least three pixels within five frames.

The Las Vegas video consists of 700 annotated frames, whose GSD is $\sim$1.0\,m and its frame rate is 30\,fps. The available video was computationally increased in resolution and frame-to-frame registered~\cite{murthy2014skysat}. Zhang et al.~\cite{zhang-corr2019b} defined two AOIs (see Fig.~\ref{fig:lasvegas}) and created out of these two AOIs two separate videos (Video 001 $\doteq$ AOI\,1, and Video 002 $\doteq$ AOI\,2). We use the same bounding box labels of moving vehicles for AOI\,1 and AOI\,2, however bounding boxes do not provide additional information to point labels~\cite{WPAFB, sommer2016survey}, because of the low visual resolution of the vehicles. We therefore use the center points of the ground truth bounding boxes.

Both satellite videos from Agadez (Fig.~\ref{fig:exampleAgadez}) and Khartoum (Fig.~\ref{fig:khartoum}) are recorded as panchromatic images, with a duration of 60\,s, 30\,fps and an image size of $2560 \times 1080$\,pixel with $\approx$95\,cm GSD.
Khartoum differs from Agadez by the traffic density. The number of training data for Agadez is therefore lower compared to Khartoum (Tab.~\ref{tab3}).
The recorded video frames are provided in 16-bit GeoTIFF format which we convert to 8-bit for the subsequent frame registration (Sec.~\ref{sec:registration}). The selected AOIs were manually labelled by characterizing the position of a vehicle by its center point. The non-overlapping AOIs are divided into training and test data (see Fig.~\ref{fig:khartoum}). The satellite's ego-motion introduces gaps at the image borders between two consecutive views. Consequently, we choose AOIs for producing nuisance-free training data to lie in the center of the video frame. AOIs for the test data were chosen along the borders and show in some cases missing pixels. Due to the severe viewpoint changes, some AOIs for testing cannot be shown in Fig.~\ref{fig:khartoum}. The reader is for these cases and for the AOIs of Agadez referred to the supplementary material.

\subsection{Registration of Video Frames}
\label{sec:registration}
The approach of Reilly et al.~\cite{reilly2010detection} is adopted for frame registration.
The method assumes structured textures for interest point occurrences which is valid for the chosen data. For two video frames, Harris corners~\cite{Harris88} are detected and then SIFT feature descriptors~\cite{lowe1999object} are computed. A Flann-based matcher~\cite{muja2009fast} compares those features and finally finds correspondences between points of both frames. Given these initial matches, MAGSAC~\cite{barath2019magsac, barath2019magsacplusplus} is used to robustly compute a homography. This pairwise registration is done between a reference frame (first video frame) and all other frames independently. Finally, all frames are warped by the homographies to the reference frame.

\begin{table}
\small
\caption{Comparison of the WPAFB 2009, Las Vegas Khartoum and Agadez datasets. Note, the number of training and test data was calculated assuming five channels. The first row (\# vehicles) gives the number of annotated vehicles selected for training and testing. The number of training samples for Las Vegas is given for AOI\,2.}
\centering
\begin{tabular}{|l|c|c|c|c|}
\hline
\textbf{} & \textbf{WPAFB} & \textbf{Las Vegas} & \textbf{Khartoum} & \textbf{Agadez} \\
\hline
\# vehicles  & > $1.7 \times 10^5$ & 80,047 & 87,274 & 18,355 \\
\hline
train samples  &  23,725 & 6,258 &  15,367  & 9,705\\
\hline
test samples & 9,321 & 4,024 & 4,925 & 2,294\\
\hline
\end{tabular}
\label{tab3}
\end{table}

\subsection{Training}
Because of the limited GPU memory, the AOIs used for training are split into small image tiles denoted as regions of objects of interest (ROOBI) with e.g. an edge size of $N$=128\,pixel (Fig.~\ref{fig:khartoum}).


Our network was trained from scratch using PyTorch - we used Adam with a learning rate of 1e-5 and a batch size of 32. An early stopping patience of 100 is employed to stop training if the validation loss is not decreasing anymore. Training and testing was performed on a single GeForce RTX 3090. We optionally applied data augmentation~\cite{perez2017effectiveness} for training. The augmentation includes adding horizontally or vertically flipped examples of all images as well as randomly translated and rotated versions.

Detections are considered as true positive if they are within a certain distance $\theta$ of a ground truth coordinate. If multiple detections are within this radius, the closest one is taken and the rest, if they do not have any other ground truth coordinates within the distance $\theta$, are marked as false positive. Any detections that are not within $\theta$ of a ground truth coordinate are also marked as false positive. Ground truth coordinates which have no detections within $\theta$ are marked as false negative. Both methods Otsu and NMS require a threshold. The Otsu threshold $\alpha_{O}$ disregards small segments of the heatmap. The NMS threshold $\alpha_{N}$ rejects all predictions with a value lower than the threshold. 


Pre-training is carried out on the three previously deccribed AOIs of the WPAFB dataset. More details about pre-training using the WPAFB data can be found in \cite{pflugfelder2020learning}.



\section{Results and Discussion}
\label{sec:results}






The main experiments were carried out on the new Khartoum and Agadez video. But in order to make the method comparable to other state-of-the-art methods, we also evaluated FoveaNet4Sat on the Las Vegas dataset.

\subsection{Las Vegas}

The FoveaNet4Sat is evaluated on the Las Vegas video with respect to the original FoveaNet and other state-of-the-art methods. We set the ROOBI side length $N$ to 200\,pixel, $\sigma=1.0$\,pixel and $\theta=8$\,pixel. The thresholds are set to $\alpha_{O}=3.5$\,pixel and $\alpha_{N}=0.35$. $c$ is set to 3.

Tab.~\ref{tab:exp6_1} illustrates that FoveaNet4Sat achieves better results than the FoveaNet. The $F_{1}$ score increases from 0.75 to 0.88 (+17.3\%) and from 0.77 to 0.86 (+11.7\%) on AOI\,1 and AOI\,2, respectively. Fine-tuning a pre-trained FoveaNet4Sat marginally changes the $F_{1}$ score in contrast to training from scratch.




\begin{table}[!htb]
	\caption{Foveanet vs. FoveaNet4Sat on Las Vegas. Training from scratch (scr.), Precision (Prec.), Recall (Rec.).}
	\begin{tabular}{|l|c|c|c||c|c|c|}
		\hline
\textbf{}&\multicolumn{3}{c||}{\textbf{AOI 1}} &\multicolumn{3}{c|}{\textbf{AOI 2}}\\
\hline
		 & \textbf{Prec.} & \textbf{Rec.} & \bm{$F_{1}$}  &  \textbf{Prec.} & \textbf{Rec.} & \bm{$F_{1}$}     \\ \hline
		\textbf{FoveaNet scr.}             & 0.71          & 0.78         & 0.75  & 0.86    & 0.71 & 0.77           \\ \hline
		\textbf{FoveaNet4Sat scr.} & 0.82          & 0.92         & 0.87  & 0.97          & 0.79         & 0.87         \\ \hline
		\textbf{FoveaNet4Sat}   & 0.85          & 0.92         & 0.88 &   0.97         & 0.78         & 0.86          \\ \hline
	\end{tabular}
	\label{tab:exp6_1}
\end{table}

The method is trained to detect the vehicle's center. However, we observe in our experiment that for vehicles of larger size, e.g. semi-trucks, buses and trains, the method frequently localises vehicle parts, such as the front of a bus. We believe that the early fusion plays here a role. The network would need to temporally relate features with a larger spatial receptive field which could be achieved by late fusion.

\begin{table}[!htb]
\caption{Our results vs. state-of-the-art on Las Vegas.}
	\begin{tabular}{|l|c|c|c||c|c|c|}
		\hline
\textbf{}&\multicolumn{3}{c||}{\textbf{AOI 1}} &\multicolumn{3}{c|}{\textbf{AOI 2}}\\
\hline
		 & \textbf{Prec.} & \textbf{Rec.}  & \bm{$F_{1}$} & \textbf{Prec.} & \textbf{Rec.}  & \bm{$F_{1}$} \\ \hline
		\textbf{GoDec \cite{zhou-icml2011}}               & 0.95          & 0.36          & 0.52    & 0.90          & 0.81          & 0.85            \\ \hline
		\textbf{RPCA-PCP \cite{candes-acm2011}}            & 0.94          & 0.41          & 0.57   & 0.90          & 0.78          & 0.84          \\ \hline
		\textbf{O-LSD \cite{zhang-corr2019b}}               & 0.65          & 0.64          & 0.64   & 0.73          & 0.90          & 0.81                   \\ \hline
		\textbf{DECOLOR \cite{zhou-tpami2013}}             & 0.77          & 0.59          & 0.67   & 0.79          & 0.81          & 0.80              \\ \hline
		\textbf{LSD \cite{liu-tip2015}}                 & 0.87          & 0.71          & 0.78  & 0.82          & 0.91          & 0.86          \\ \hline
		\textbf{E-LSD \cite{zhang-corr2019a}}               & 0.85          & 0.79          & 0.82  & 0.80          & 0.94          & 0.86               \\ \hline
		\textbf{FoveaNet \cite{pflugfelder2020learning}}  &   0.86        &   0.82       &  0.84 &           &           &             \\ \hline
		\textbf{FoveaNet4Sat} & 0.82          & 0.92          & 0.87      & 0.97          & 0.79         & 0.87      \\ \hline
	\end{tabular}
	\label{tab:sota6_1}
\end{table}

Results in comparison to state-of-the-art methods are provided in Tab.~\ref{tab:sota6_1}. In order to stay compliant with~\cite{zhang-corr2019a} and~\cite{zhang-corr2019b} the first 200 frames of both AOIs are skipped which leads to slightly different results than those given in Tab.~\ref{tab:exp6_1}. It is shown that FoveaNet4Sat outperforms the E-LSD~\cite{zhang-corr2019a} on AOI 1. The $F_{1}$ score increases from 0.82 to 0.87 (+6.4\%). The evaluation of AOI\,2 shows that the performance of FoveaNet4Sat is comparable to the state-of-the-art methods~\cite{zhang-corr2019a, liu-tip2015} in terms of $F_{1}$ score. A comparison of LSD~\cite{liu-tip2015} with Foveanet4Sat shows that precision improves from 0.82 to 0.97 while the recall decreases from 0.91 to 0.79. We think this is due the traffic junction in the lower right corner of AOI\,2 (Fig.~\ref{fig:lasvegas} where nearly halting cars appear. In contrast, AOI\,1 does not show a comparable pattern of motion.



\subsection{Khartoum}

We carried out two experiments for this dataset. For this, we set the ROOBI side length $N$ to 128\,pixel, $\sigma=1.0$\,pixel and $\theta=4$\,pixel. The thresholds are set to $\alpha_{O}=3.5$\,pixel and $\alpha_{N}=0.40$. 

\begin{figure}[b]
\centering
\includegraphics[width=0.48\textwidth]{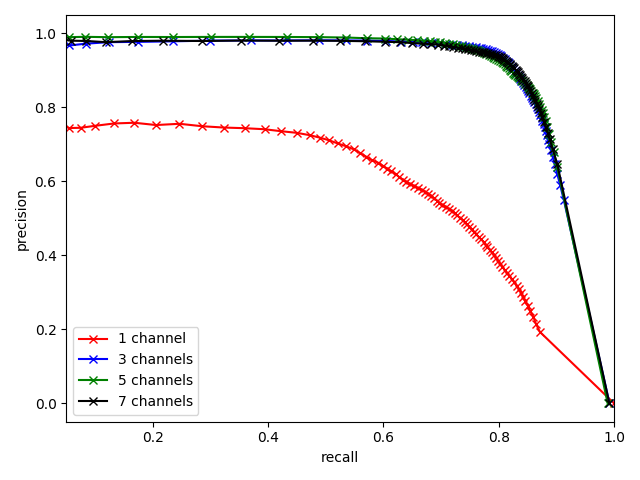}
\caption{Precision-recall curve for one, three, five and seven channels as input to the FoveaNet4Sat.}
\label{fig:resultskhartoum_channels}
\end{figure}

{\bf Impact of spatiotemporal features:} The first experiment compares spatial against spatiotemporal features and confirms on new video data with more complex traffic patterns the power of spatiotemporal data for tiny object detection. 
For this, the Foveanet4Sat is trained with various numbers of channels and from scratch with 14,500 training samples. The impact of different numbers of channels on the detection is presented by precision-recall curves (Fig.~\ref{fig:resultskhartoum_channels}).
It is obvious that the network benefits from learning spatiotemporal features. Sole spatial features in a single frame would not allow a detector to exceed a precision of 0.5 for a recall of 0.8, whereas three frames or more give a precision larger than 0.9. It is interesting to see that five frames marginally improve this result. The accuracy of the model does not increase any more if the number of channels is further increased to seven frames. One hypothesis is that already three frames give the necessary and sufficient spatiotemporal information to reliably detect vehicles. This additional information might be carried by elongated blobs in the space-time video, whereas a single blob (4-10 pixel) of a vehicle in the single image does not contain enough information to discriminate vehicle from clutter. An analysis of the learnt spatiotemporal features is necessary to gather more evidence, e.g. by the visualisation method in \cite{springenberg-iclrw2015}.


In a second experiment, the model is trained with eight different configurations as summarized in Tab.~\ref{configs_foveanet} and with a varying number of training samples between 5 and 15,300. We additionally perform data augmentation for the case of 15,300 samples (15,300+).
The training samples were randomly selected. The training for each configuration was carried out three times to better understand the influence of the individual training on the result. The results of the conducted experiment is summarised in Tab.~\ref{tab:training}. 

\begin{figure}
\centering
\includegraphics[width=0.48\textwidth]{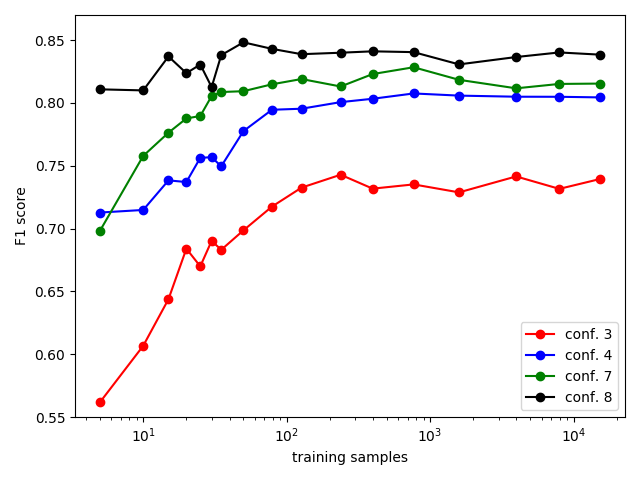}
\caption{$F_1$ score vs. the number of training samples for all configurations including the fine-tuning.}
\label{fig:resultskhartoum}
\end{figure}

{\bf Does FoveaNet4Sat do better than Foveanet?}
For example, by comparing Foveanet and FoveaNet4Sat when trained from scratch with the same 15,300 samples, FoveaNet4Sat receives significantly better results on the test data than Foveanet. The $F_1$ score increases from 0.6 (config.\,1) to 0.83 (config.\,5) which is an improvement of +38.3\%! The variation of the three independent training rounds show negligible variations (standard deviation is near zero). The result even improves slightly to 0.86 by applying data augmentation.
Moreover the results of the experiment indicate that FoveaNet4Sat performs constantly better than Foveanet over all configurations with fine-tuning, even when the number of training samples is reduced (Fig.~\ref{fig:resultskhartoum}).

\begin{figure*}
	\centering
	\begin{subfigure}[t]{0.48\textwidth}
		\centering
		\includegraphics[width=\textwidth]{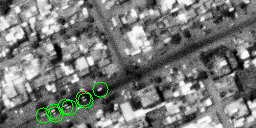}
                \caption{}\label{fig:k_aoi_0}
	\end{subfigure}
	\hspace*{\fill}
	\begin{subfigure}[t]{0.48\textwidth}
		\centering
		\includegraphics[width=\textwidth]{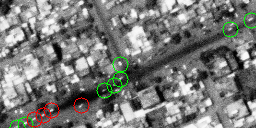}
                \caption{}\label{fig:k_aoi_1}
	\end{subfigure}
	\vspace*{\fill}
        \begin{subfigure}[t]{0.435\textwidth}
		\centering
		\includegraphics[width=\textwidth]{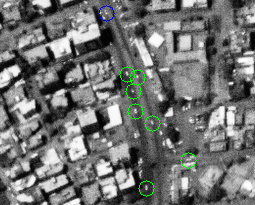}
                \caption{}\label{fig:k_aoi_2}
	\end{subfigure}
	\hspace*{\fill}
	\begin{subfigure}[t]{0.545\textwidth}
		\centering
		\includegraphics[width=\textwidth]{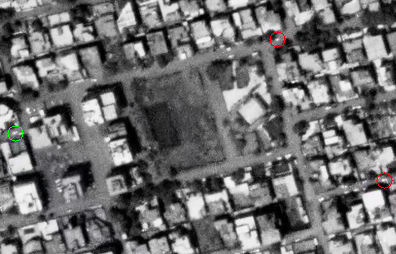}
                \caption{}\label{fig:k_aoi_3}
	\end{subfigure}
	\vspace*{\fill}
        \begin{subfigure}[t]{0.485\textwidth}
		\centering
		\includegraphics[width=\textwidth]{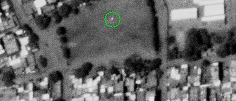}
                \caption{}\label{fig:k_aoi_4}
	\end{subfigure}
	\hspace*{\fill}
	\begin{subfigure}[t]{0.475\textwidth}
		\centering
		\includegraphics[width=\textwidth]{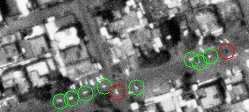}
                \caption{}\label{fig:k_aoi_5}
	\end{subfigure}

	\caption{Qualitative results of Khartoum (true positive: green, false positive: blue, false negative: red).} \label{fig:resultskhartoum_pics} 
\end{figure*}

{\bf Qualitative results:} Fig.~\ref{fig:resultskhartoum_pics} shows examples of the detector's output for selected scenes. The videos to all examples can be found in the supplemental material. Although vehicles are very proximate, the method is able to correctly detect all vehicles (Fig.~\ref{fig:k_aoi_0}). For nearly the same scenery, the method is suddenly unable to detect the vehicles (Fig.~\ref{fig:k_aoi_1}). We believe that the cause of these false negatives is the slow vehicle movement which is better seen in the corresponding video. Although a false positive, the detected vehicle in Fig.~\ref{fig:k_aoi_2} seems to be correct. This emphasises the difficulty to manually label the data. Fig.~\ref{fig:k_aoi_3} depicts two false negatives and a correctly detected vehicle. We believe that these failure cases emerge due to the strong clutter in the vicinity of the vehicles which interferes with the spatiotemporal pattern of the vehicles. The method is also able to detect a vehicle off-road (Fig.~\ref{fig:k_aoi_4}). Finally, Fig.~\ref{fig:k_aoi_5} depicts a failure case where a vehicle with dark appearance is missed. Most vehicles in the training data are represented as bright, white blobs. More qualitative results can be found in the supplementary material.

\begin{table}
\small
\caption{Configurations for training.}
\centering
\begin{tabular}{|c|c|c|c|}
\hline
\textbf{config.}  & \textbf{model} & \textbf{channels} & \textbf{training} \\
\hline
1 &  FoveaNet & 3 & scratch  \\
\hline
2 &  FoveaNet & 5 & scratch  \\
\hline
3 &  FoveaNet & 3 & fine-tuned  \\
\hline
4   & FoveaNet & 5 & fine-tuned  \\
\hline
5  & FoveaNet4Sat & 3 & scratch  \\
\hline
6  & FoveaNet4Sat & 5 & scratch  \\
\hline
7 & FoveaNet4Sat & 3 & fine-tuned  \\
\hline
8 & FoveaNet4Sat & 5 & fine-tuned  \\
\hline
\end{tabular}
\label{configs_foveanet}
\end{table}

{\bf How many labelled samples are necessary?} 
This is an important question, as sufficient accuracy, e.g. $F_1 > 0.8$, on new video with low annotation effort is needed for the practicability of the detector. 
The second experiment clearly shows that e.g. a training with the Foveanet (config.\,3) does not allow this practicability by a large margin. By considering five channels and a training with fine-tuning of FoveaNet4Sat (config.\,8) the results give a $F_1$ score of $0.81\pm 0.02$ (Tab.~\ref{tab:training}). It is remarkable that this performance is achieved on the test data with only five training samples, i.e. five images (mid images of a training sample) need to be manually annotated to get this result. Interestingly, much more manually annotated images marginally increase this result. For example, additional 7,959 annotated images only increase the $F_1$ score by only 3.7\,\%! The same training setup with three channels reaches $F_1 = 0.7\pm 0.07$ which is significantly lower. Our hypothesis is that two additional channels better discriminate spatiotemporal patterns of moving vehicles from clutter. A closer look on the $F_1$ score shows that the drop is caused by the recall (config.\,8 $0.7\pm 0.035 \to$ config.\,7 $0.55\pm 0.09$) and not by the precision (config.\,8 $0.97\pm 0.01 \to$ config.\,7 $0.96\pm 0.01$) which shows evidence for this hypothesis. These patterns emerge in video by the temporal coherence assumption. A deeper analysis of these learnt patterns would be necessary.

Fig.~\ref{fig:resultskhartoum} further shows that FoveaNet4Sat outperforms Foveanet in all configurations especially for training the models with a low number of samples. Configurations 1, 2, 5, 6 (training from scratch) are not shown. For a low number of training samples training from scratch performs significantly worse than training with fine-tuning.

\begin{SCtable*}
\small
\centering
	\caption{Comparison between the different configurations (config.) and the number of training samples (\# data) from the Khartoum video. Each entry shows a $F_1$ score's statistic of three training rounds by the mean and the standard deviation (in brackets).}

	\begin{tabular}{|c|c|c|c|c||c|}
		\hline
\textbf{\# data}  &   5         &     10      & 7,964        &   15,300     &  15,300+  \\ \hline \hline
\textbf{config.}&\multicolumn{5}{c|}{\textbf{\bm{${F_{1}}$} scores' statistics}} \\
\hline
\textbf{1}  & 0.01 (0.00) & 0.01 (0.01) & 0.58 (0.02) & 0.60 (0.01) &  0.81 (0.00)\\ \hline
\textbf{2}  & 0.01 (0.00) & 0.02 (0.01) & 0.72 (0.01) & 0.74 (0.01) &  0.82 (0.00)\\ \hline
\textbf{3}  & 0.56 (0.03) & 0.61 (0.02) & 0.73 (0.01) & 0.74 (0.01) &  0.81 (0.01)\\ \hline
\textbf{4}  & 0.71 (0.01) & 0.71 (0.02) & 0.80 (0.01) & 0.80 (0.00) &  0.83 (0.00)\\ \hline
\textbf{5}     & 0.14 (0.15) & 0.24 (0.34) & 0.83 (0.00) & 0.83 (0.01) &  0.86 (0.00)\\ \hline
\textbf{6}     & 0.24 (0.28) & 0.26 (0.37) & 0.83 (0.01) & 0.84 (0.01) &  0.86 (0.00)\\ \hline
\textbf{7}     & 0.70 (0.07) & 0.76 (0.02) & 0.82 (0.00) & 0.82 (0.00) &  0.86 (0.00)\\ \hline
\textbf{8}     & 0.81 (0.02) & 0.81 (0.03) & 0.84 (0.00) & 0.84 (0.00) &  0.86 (0.00)\\ \hline


\end{tabular}
\label{tab:training}
\end{SCtable*}

\subsection{Agadez}
This video shows compared to Las Vegas and Khartoum a small, rural town in the Sahel zone. The cityscape is dominated by unpaved roads and small houses built from clay. The street scene is dominated by motorcycles and pedestrians. The occurrence of vehicles is rather low.
FoveaNet4Sat is trained as given by config.\,8 with the annotated training samples (Tab.~\ref{tab3}). Data augmentation was additionally applied. All the training parameters are set to the same values as for the training of the Khartoum video.
The detector achieves a precision of 0.96 and a recall of 0.70 ($F_1 = 0.81$) on the test data (Tab.~\ref{tab3}). This result is slightly worse compared to the result of Khartoum (Tab.~\ref{tab:training}, config.\,8, 7,964 training samples).
Additional to traffic density which yields less training data, a lower image contrast and increased noise caused by atmospheric disturbances during the acquisition makes the Agadez video more challenging for our method than the Khartoum video.

\begin{figure*}
	\centering
	\begin{subfigure}[t]{0.42\textwidth}
		\centering
		\includegraphics[width=\textwidth]{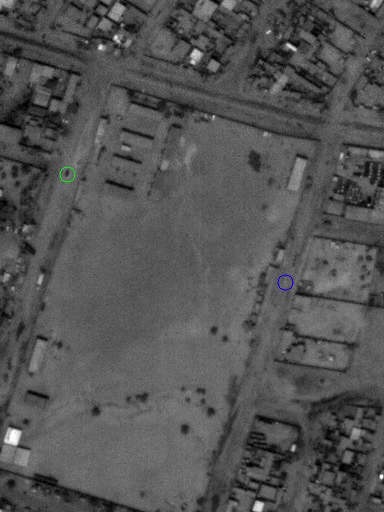}
                \caption{}\label{fig:ag_aoi_5}
	\end{subfigure}
	\begin{subfigure}[t]{0.56\textwidth}
		\centering
		\includegraphics[width=\textwidth]{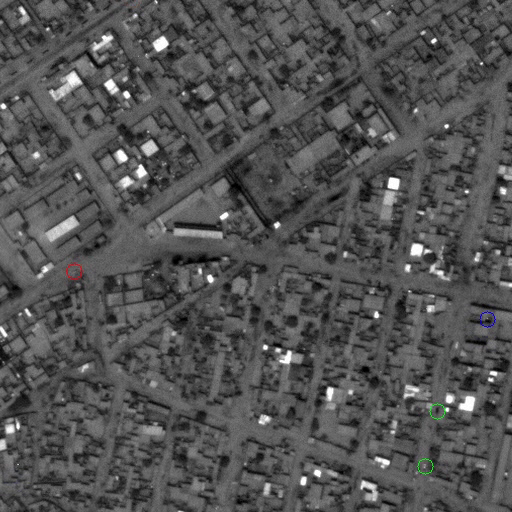}
                \caption{}\label{fig:ag_aoi_4}
	\end{subfigure}

	\caption{Qualitative results of Agadez (true positive: green, false positive: blue, false negative: red).} \label{fig:results_agadez} 
\end{figure*}

Fig.~\ref{fig:results_agadez} shows that the method detects in some cases the even smaller motorcycles which are not labelled as vehicles and which are counted as false positives (Fig.~\ref{fig:ag_aoi_5}). We believe due to the lacking spatial information, the detector is not able to distinguish cars from motorcycles. Fig.~\ref{fig:ag_aoi_4} depicts a false negative, we believe due to the vehicle's similarity with the background (low contrast). 
Interestingly, this example also shows a false positive which we believe is neither a street vehicle nor an artifact of the acquisition. The reader is referred to the corresponding video for a better impression.


\section{Conclusion}
\label{sec:conclusion}
The work presents a vehicle detector for satellite video based on deep learning. The proposed spatiotemporal model of a compact $3 \times 3$ convolutional, neural network and the new post-processing improves our first attempt~\cite{pflugfelder2020learning} and the state-of-the-art on the Las Vegas benchmark. It turned out that model changes such as neglecting max-pooling and using leaky ReLUs as well as replacing Otsu thresholding with NMS improve the detection significantly for the case of proximate vehicles.

The work also brings new empirical results on two new satellite videos which we have acquired and labelled and which we plan to make available to the community. The results show that spatiotemporal features and spatiotemporal networks are able to successfully exploit the temporal consistency of moving vehicles. The new video data acknowledges again this approach~\cite{pflugfelder2020learning}.
The work reveals that a reliable detection on previously unseen video can be accomplished with a few annotated training samples. Our experiment shows that fine-tuning with five training samples containing five manually labelled satellite images yield a reliable detector ($F_1 = 0.81 \pm 0.02$).
Supervised learning usually assumes large annotated data which is difficult to achieve for satellite video. The results show that the deep learning approach of pre-training on a large WAMI dataset and then transfer learning (fine-tuning) on a few manually labelled satellite video frames might promise a solution for vehicle detection in satellite imagery.

Future work will be necessary to understand the nature of these spatiotemporal features and how they are formed. A suggestion is to visualise the activated input neurons which is now easily possible as FoveaNet4Sat does not contain a pooling layer and follows in this respect earlier work that shows such a visualisation technique~\cite{springenberg-iclrw2015} for 2D images. Such visualisation will also let us better understand the potential correlation between recall, the complexity of motion patterns and the network's number of channels.
Finally, a theoretical analysis of 2D spatiotemporal networks and their comparison to 3D networks could give new insight to their capabilities and limitations.



%

\appendices


\section*{Acknowledgment}
This project has received funding from the European Union’s Horizon 2020 research and innovation programme under the Marie Skłodowska-Curie grant agreement No 899987, the Secure Societies grant agreement No 787021 and from the Austrian Security Research Programme KIRAS of the Austrian Research Promotion Agency (FFG) under the grant agreement No 867030. The imagery and video used in this work is in courtesy of the U.S. air force research laboratory sensors directorate layered sensing exploitation division and Planet Inc. respectively. All images of Fig.~\ref{fig:intro:challenge}, \ref{fig:exampleAgadez},\ref{fig:NMSversusOtsu}, \ref{fig:khartoum}, \ref{fig:resultskhartoum_pics}, \ref{fig:results_agadez} are copyright protected (\textcopyright 2020, Planet Labs Inc). All Rights Reserved. The image in Fig.~\ref{fig:lasvegas} was provided by Skybox Inc. under the creative common CC-BY-NC-SA\footnote{\url{https://creativecommons.org/licenses/by-nc-sa/4.0/legalcode}}. We thank Junpeng Zhang for providing us the annotation of the Las Vegas video and Rohan Agrawal, Cogito Inc. for his patience and support during the annotation of the Khartoum and Agadez videos. We are thankful to Marcus Apel and Regina Kozyra, Planet Labs Inc. for their support and advice with the data.

\ifCLASSOPTIONcaptionsoff
  \newpage
\fi

\section{Supplemental Material}
We show more intermediate results in this supplementary material to give the reader a better understanding of our method. In Section~\ref{sec:las-vegas} we provide additional results for our method for the Las Vegas video. In Section~\ref{sec:khartoum}, we provide even more detailed information on the quantitative results and also briefly explain the videos with the detection results. In Section~\ref{sec:agadez}, we show the selected training and evaluation AOIs and provide the consecutive videos. All videos can be downloaded here: \url{https://cloud.vision.in.tum.de/s/q8AL3JFNZxcNWfX}.

\subsection{Experiments and results of the Las Vegas video}
\label{sec:las-vegas}

\begin{figure*}
	\centering
        \begin{minipage}{.55\textwidth}
	\begin{subfigure}[t]{0.48\columnwidth}
		\centering
		\includegraphics[width=.8\textwidth]{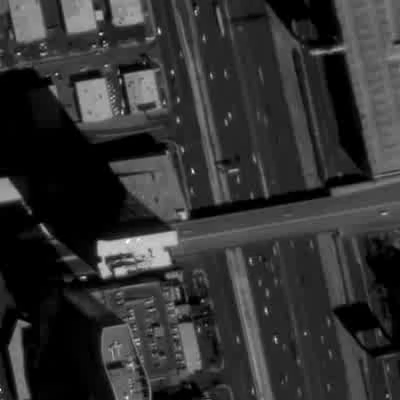}
		\caption{AOI\,1} \label{fig:vegas_res_01}
	\end{subfigure}
	\begin{subfigure}[t]{0.48\columnwidth}
		\centering
		\includegraphics[width=.7\textwidth]{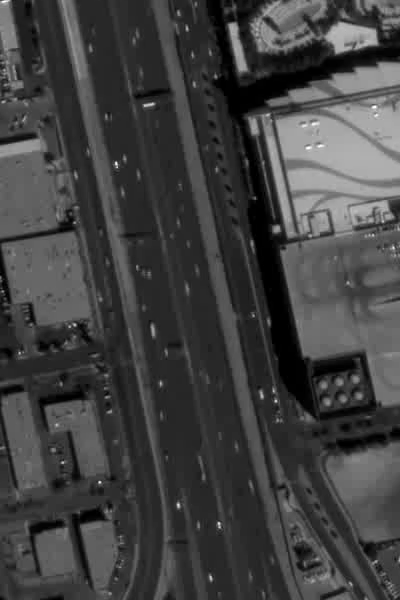}
		\caption{AOI\,2} \label{fig:vegas_02}
	\end{subfigure}
	\caption{Enlarged depiction of both Las Vegas satellite video AOIs.} \label{fig:vegas_large} 
\end{minipage}%
\begin{minipage}{.4\textwidth}
 \centering
 \includegraphics[width=0.8\textwidth]{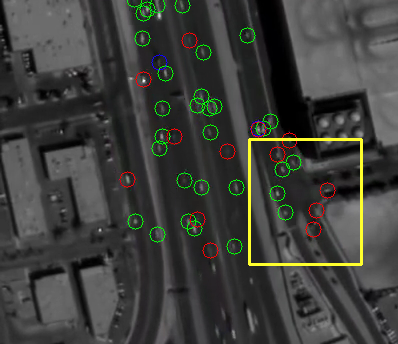}
 \caption{A traffic junction with slow and halting vehicles (yellow rectangle) is challenging for FoveaNet4Sat.} \label{fig:junction} 
\end{minipage}%
\end{figure*}

\begin{figure*}
	\centering
	\begin{subfigure}[t]{0.16\columnwidth}
		\centering
		\includegraphics[width=\textwidth]{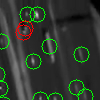}
		\caption{} \label{fig_supple:vegas_res_01}
	\end{subfigure}
	\begin{subfigure}[t]{0.16\columnwidth}
		\centering
		\includegraphics[width=\textwidth]{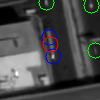}
		\caption{} \label{fig:vegas_res_02}
	\end{subfigure}
	\begin{subfigure}[t]{0.16\columnwidth}
		\centering
		\includegraphics[width=\textwidth]{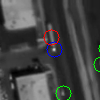}
		\caption{} \label{fig:vegas_res_03}
	\end{subfigure}
	\begin{subfigure}[t]{0.16\columnwidth}
		\centering
		\includegraphics[width=\textwidth]{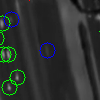}
                 \caption{} \label{fig:vegas_res_04}
	\end{subfigure}
	\begin{subfigure}[t]{0.16\columnwidth}
		\centering
		\includegraphics[width=\textwidth]{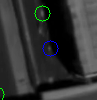}
                \caption{} \label{fig:vegas_res_05}
	\end{subfigure}
	\begin{subfigure}[t]{0.16\columnwidth}
		\centering
		\includegraphics[width=\textwidth]{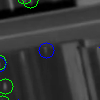}
                \caption{} \label{fig:vegas_res_06}
	\end{subfigure}

	\caption{Difficulties for the detector and human annotator.} \label{fig:large} 
\end{figure*}

To further illustrate the Foveanet4Sat's performance on the Las Vegas video, we provide two supplemental videos showing our results (Fig.~\ref{fig:vegas_large}). The two videos for both AOIs are named \verb$vegas_AOI1.mp4$ and \verb$vegas_AOI2.mp4$ which are available at the given weblink under the folder \verb$lasvegas$. Notes on the results:
\begin{itemize}
\item As mentioned in the article, Las Vegas AOI\,2 has in the bottom right a junction (Fig.~\ref{fig:junction}) where stop-and-go traffic can be observed. Certainly, this is a challenge for detectors. 
\item We mention in the article the problem that vehicles of atypical size, e.g. semi-trucks, buses and trains are not always correctly detected. In the following, we show images of example results where the method instead of the vehicle's center (\ref{fig:vegas_res_01}) detects,  e.g. the front or rear of a bus or semi-truck (Fig.~\ref{fig:vegas_res_02}, \ref{fig:vegas_res_03}). 
\item We mention the challenge of the manual annotation of the videos. The method was able to detect a vehicle correctly (this was ex-post verified) for some cases. This increases the number of false positives incorrectly. Fig.~\ref{fig:vegas_res_04}, \ref{fig:vegas_res_05}, \ref{fig:vegas_res_06} show examples.
\end{itemize}


\subsection{Experiments and results of the Khartoum video}
\label{sec:khartoum}

\begin{figure*}
\centering
\includegraphics[width=0.95\textwidth]{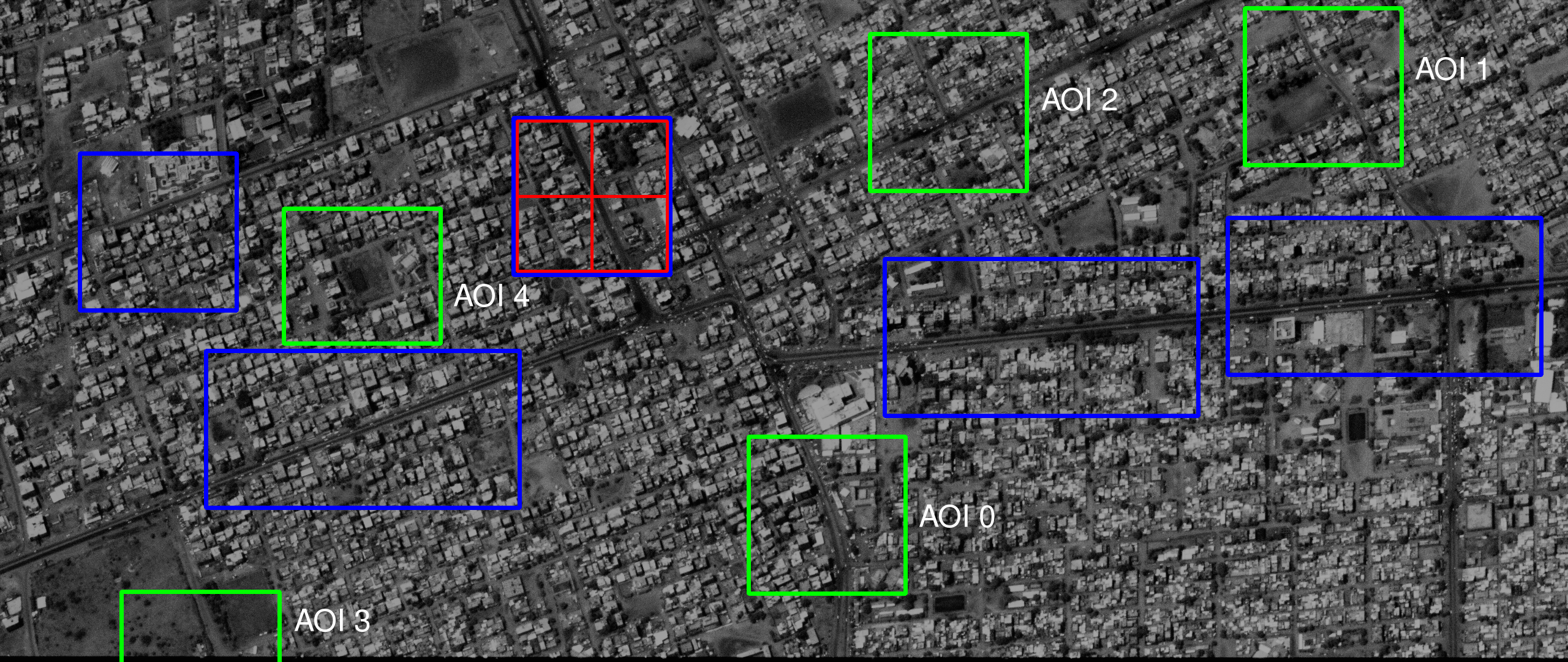}
\caption{First Khartoum video frame. AOIs used for training and evaluation are marked by blue and green rectangles. This illustration additionally shows the AOI numbering of the test data.}
\label{fig:khartoum_aois}
\end{figure*}

Tab.~\ref{tab:training2} is an extended version of our results in the paper and shows precision and recall explicitly for experiment two. Fig.~\ref{fig:khartoum_aois} shows all selected AOIs for the Khartoum video. The AOIs used for the evaluation are numbered consecutively so that the reader can associate the video with the corresponding AOI.
The result videos illustrating the detections of our method for all test AOIs (Fig.~\ref{fig:khartoum_aois} ) can be found at the weblink under the folder \verb$khartoum$:

\begin{table*}
\small
\centering
\caption{Comparison between the different configurations (config.) and the number of training samples (\# data) from the Khartoum video. Each entry shows $F_1$ and the precision (Prec.), recall (Rec.) score's statistics of three training rounds by the mean and the standard deviation (in brackets).}
\begin{tabular}{|c|c|c|c||c|c|c||c|c|c|}
\hline
\textbf{\# data}  &  \multicolumn{3}{c||} 5         &      \multicolumn{3}{c||}{10}      & \multicolumn{3}{c|}{7,964}          \\ \hline 
\textbf{config.} & Prec. & Rec. & \bm{${F_{1}}$}   & Prec. & Rec. & \bm{${F_{1}}$} & Prec. & Rec. & \bm{${F_{1}}$}  \\
\hline
\textbf{1}  & 0.01 (0.00)& 0.02 (0.00)& 0.01 (0.00)& 
              0.02 (0.01)& 0.01	(0.01)& 0.01 (0.01)&
   	      0.55 (0.03)& 0.61	(0.01)& 0.58 (0.02)\\ \hline
\textbf{2}  & 0.01 (0.00)& 0.01	(0.01)& 0.01 (0.00)& 
0.02 (0.01)& 0.02 (0.01) & 0.02 (0.01) &
0.75 (0.04)& 0.68 (0.01)&  0.72 (0.01) \\ \hline
\textbf{3} & 0.68 (0.14)& 0.50 (0.06)& 0.56 (0.03) &
0.73 (0.09)& 0.54 (0.08)& 0.61 (0.02) & 
0.81 (0.02)& 0.67 (0.00)& 0.73 (0.01) \\ \hline
\textbf{4}  & 0.83 (0.09)& 0.63 (0.04)& 0.71 (0.01)& 
0.76 (0.05)& 0.68 (0.01)& 0.71 (0.02) &
 0.91 (0.00)& 0.72 (0.01)& 0.80 (0.01) \\ \hline
\textbf{5}  & 0.62 (0.44)& 0.08 (0.10)& 0.14 (0.15) &
0.33 (0.46)& 0.19 (0.27)&  0.24 (0.34) &
0.91 (0.01)& 0.77 (0.01)&  0.83 (0.00)\\ \hline
\textbf{6} & 0.63 (0.45)& 0.17 (0.21)& 0.24 (0.28)& 
0.33 (0.46)& 0.22 (0.30) & 0.26 (0.37)& 
0.91 (0.01)& 0.77 (0.01) & 0.83 (0.01)\\ \hline
\textbf{7} &0.96 (0.01)& 0.55 (0.09) & 0.70 (0.07) &
0.95 (0.01)& 0.63 (0.02) & 0.76 (0.02)& 
0.87 (0.01)& 0.76 (0.00) & 0.82 (0.00)\\ \hline
\textbf{8} &0.97 (0.00)& 0.70 (0.04) & 0.81 (0.02) &
 0.97 (0.00)& 0.70 (0.02)& 0.81 (0.03) &
0.92 (0.00)& 0.78 (0.01)&  0.84 (0.00) \\ \hline 	
\end{tabular}

\bigskip
\begin{tabular}{|c|c|c|c||c|c|c|}
\hline
\textbf{\# data}  &  \multicolumn{3}{c||}{15,300}         &      \multicolumn{3}{c|}{15,300+}         \\ \hline 
\textbf{config.} & Prec. & Rec. & \bm{${F_{1}}$}   & Prec. & Rec. & \bm{${F_{1}}$}   \\
\hline
\textbf{1}  & 0.57 (0.02)& 0.63 (0.00)& 0.60 (0.01)& 
   	      0.92 (0.00)& 0.73	(0.00)& 0.81 (0.00)\\ \hline
\textbf{2}  & 0.81 (0.03)& 0.68	(0.01)& 0.74 (0.01)& 
0.93 (0.00)& 0.74 (0.01)&  0.82 (0.00) \\ \hline
\textbf{3} & 0.83 (0.03)& 0.67 (0.01)& 0.74 (0.01) &
0.91 (0.02)& 0.72 (0.01)& 0.81 (0.01) \\ \hline
\textbf{4}  & 0.91 (0.01)& 0.72 (0.00)& 0.80 (0.00)& 
 0.92 (0.01)& 0.75 (0.00)& 0.83 (0.00) \\ \hline
\textbf{5}  & 0.90 (0.02)& 0.77 (0.00)& 0.83 (0.01) &
0.93 (0.00)& 0.80 (0.00)&  0.86 (0.00)\\ \hline
\textbf{6} & 0.92 (0.01)& 0.77 (0.01)& 0.84 (0.01)& 
0.92 (0.00)& 0.80 (0.00) & 0.86 (0.00)\\ \hline
\textbf{7} &0.89 (0.01)& 0.75 (0.00) & 0.82 (0.00) &
0.92 (0.01)& 0.80 (0.00) & 0.86 (0.00)\\ \hline
\textbf{8} &0.92 (0.01)& 0.77 (0.01) & 0.84 (0.00) &
0.92 (0.00)& 0.81 (0.01)&  0.86 (0.00) \\ \hline 
\end{tabular}
\label{tab:training2}
\end{table*}

\begin{multicols}{2}
\small
\begin{itemize}
\item  \verb$khartoum_AOI_0.mp4$
\item \verb$khartoum_AOI_0_v1.mp4$ 

\item  \verb$khartoum_AOI_1.mp4$ 

\item  \verb$khartoum_AOI_2.mp4$ 

\item  \verb$khartoum_AOI_2_v1.mp4$ 

\item  \verb$khartoum_AOI_2_v2.mp4$ 
\end{itemize}

\columnbreak

\begin{itemize}

\item  \verb$khartoum_AOI_3.mp4$ 

\item  \verb$khartoum_AOI_4.mp4$ 

\item  \verb$khartoum_AOI_5.mp4$ 

\end{itemize}
\end{multicols}


\subsection{Experiments and results of the Agadez video}
\label{sec:agadez}

\begin{figure*}
\centering
\includegraphics[width=0.95\textwidth]{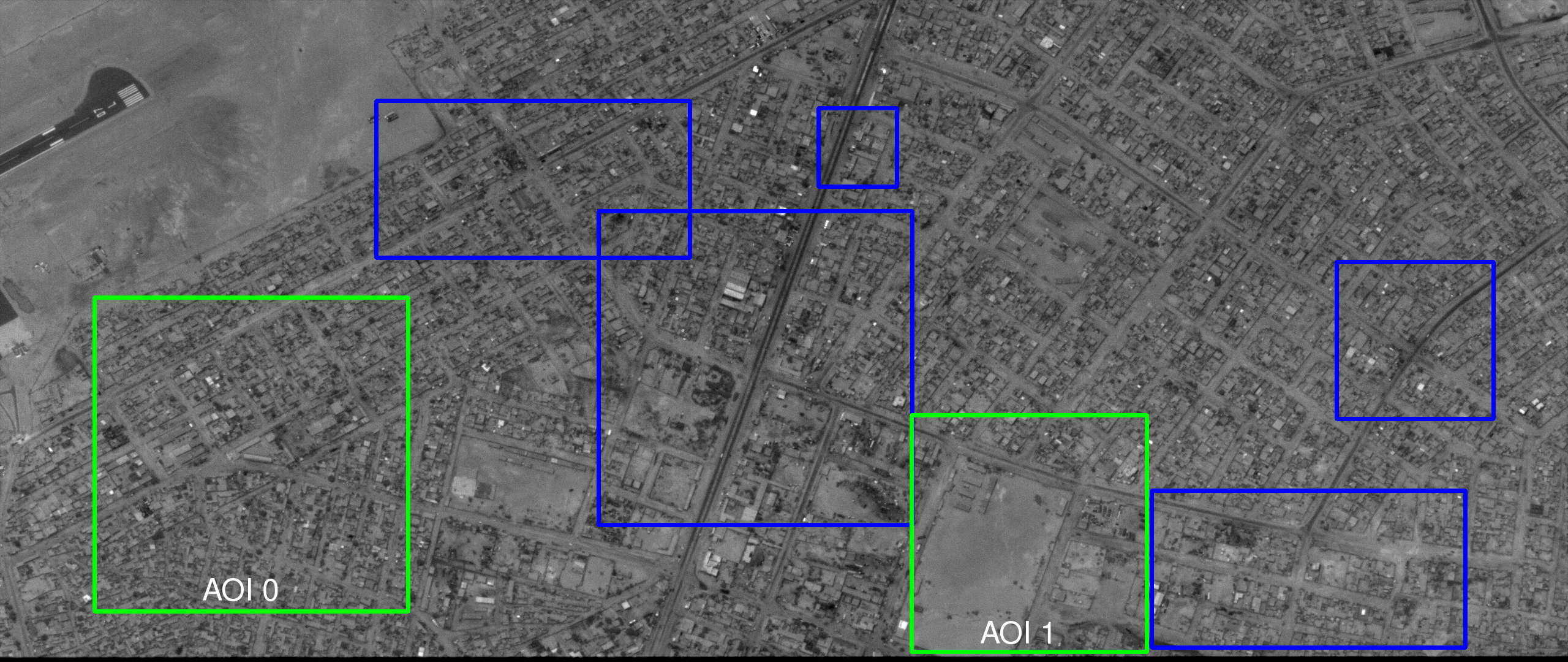}
\caption{First video frame of the Agadez video. AOIs used for training (blue) and evaluation (green) are shown as rectangles. This illustration additionally shows the AOI numbering of the test data.}
\label{fig:agadez_aois}
\end{figure*}

Fig.~\ref{fig:agadez_aois} shows all selected AOIs for the Agadez video. The AOIs used for the evaluation are numbered consecutively so that the reader can associate the video with the corresponding AOI. The result videos (\verb$aga_aoi0.mp4$ and \verb$aga_aoi1.mp4$) illustrating the detections of our method for both test AOIs (Fig.~\ref{fig:agadez_aois} ) can be found at the weblink under the folder \verb$agadez$.

%



{\small
\bibliographystyle{ieee_fullname}
\bibliography{egbib}

\begin{thebibliography}{10}\itemsep=-1pt

\bibitem{WPAFB}
AFRL.
\newblock Wright-patterson air force base (wpafb) dataset.
\newblock \url{https://www.sdms.afrl.af.mil/index.php?collection=wpafb2009},
  2009.

\bibitem{ahmadi-ijrs2019}
S.~A. Ahmadi, A. Ghorbanian, and A. Mohammadzadeh.
\newblock Moving vehicle detection, tracking and traffic parameter estimation
  from a satellite video: a perspective on a smarter city.
\newblock {\em IJRS}, 40(22):8379--8394, 2019.

\bibitem{AL-SHAKARJI_2019_CVPR_Workshops}
N. Al-Shakarji, F. Bunyak, H. Aliakbarpour, G. Seetharaman, and K. Palaniappan.
\newblock Multi-cue vehicle detection for semantic video compression in
  georegistered aerial videos.
\newblock In {\em CVPR}, June 2019.

\bibitem{ao-tip2019}
W. {Ao}, Y. {Fu}, X. {Hou}, and F. {Xu}.
\newblock Needles in a haystack: Tracking city-scale moving vehicles from
  continuously moving satellite.
\newblock {\em IEEE TIP}, preprint:1--1, 2019.

\bibitem{ao-corr2018}
Wei Ao, Yanwei Fu, and Feng Xu.
\newblock Detecting tiny moving vehicles in satellite videos.
\newblock {\em CoRR}, abs/1807.01864, 2018.

\bibitem{Baccouche_2011_HBU}
Moez Baccouche, Franck Mamalet, Christian Wolf, Christophe Garcia, and Atilla
  Baskurt.
\newblock Sequential deep learning for human action recognition.
\newblock In Albert~Ali Salah and Bruno Lepri, editors, {\em Human Behavior
  Understanding}, pages 29--39, Berlin, Heidelberg, 2011. Springer Berlin
  Heidelberg.

\bibitem{Badrinarayanan-tpami2017}
Vijay Badrinarayanan, Alex Kendall, and Roberto Cipolla.
\newblock Segnet: A deep convolutional encoder-decoder architecture for image
  segmentation.
\newblock {\em IEEE Transactions on Pattern Analysis and Machine Intelligence},
  39(12):2481--2495, 2017.

\bibitem{barath2019magsac}
Daniel Barath, Jiri Matas, and Jana Noskova.
\newblock {MAGSAC}: marginalizing sample consensus.
\newblock In {\em Conference on Computer Vision and Pattern Recognition}, 2019.

\bibitem{barath2019magsacplusplus}
Daniel Barath, Jana Noskova, Maksym Ivashechkin, and Jiri Matas.
\newblock {MAGSAC}++, a fast, reliable and accurate robust estimator.
\newblock In {\em Conference on Computer Vision and Pattern Recognition}, 2020.

\bibitem{benenson-cvpr2013}
R. {Benenson}, M. {Mathias}, T. {Tuytelaars}, and L. {Van Gool}.
\newblock Seeking the strongest rigid detector.
\newblock In {\em CVPR}, June 2013.

\bibitem{bodla2017soft}
Navaneeth Bodla, Bharat Singh, Rama Chellappa, and Larry~S Davis.
\newblock Soft-nms--improving object detection with one line of code.
\newblock In {\em Proceedings of the IEEE international conference on computer
  vision}, pages 5561--5569, 2017.

\bibitem{candes-acm2011}
E.~J. Cand\`{e}s, X. Li, Y. Ma, and J. Wright.
\newblock Robust principal component analysis?
\newblock {\em J. ACM}, 58(3):11:1--11:37, June 2011.

\bibitem{chen-access2020}
Renxi Chen, Xinhui Li, and Shengyang Li.
\newblock A lightweight cnn model for refining moving vehicle detection from
  satellite videos.
\newblock {\em IEEE Access}, 8:221897--221917, 2020.

\bibitem{chen-spie2021}
Renxi Chen, Lu Wang, Mengli Zhang, Xinhui Li, and Shengyang Li.
\newblock {Using background model and shallow convolutional neural network to
  detect moving vehicles from satellite videos}.
\newblock In Junhong Su, Junhao Chu, Qifeng Yu, and Huilin Jiang, editors, {\em
  Seventh Symposium on Novel Photoelectronic Detection Technology and
  Applications}, volume 11763, pages 1021 -- 1030. International Society for
  Optics and Photonics, SPIE, 2021.

\bibitem{Cohenour2015}
C. {Cohenour}, F. {Graas}, R. {Price}, and T. {Rovito}.
\newblock Camera models for the {W}right {P}atterson {A}ir {F}orce {B}ase
  {(WPAFB)} 2009 wide-area motion imagery data set.
\newblock {\em IEEE Aerospace and Electronic Systems Magazine}, 30:4--15, 06
  2015.

\bibitem{ding-jprs2018}
P. Ding, Y. Zhang, W.-J. Deng, P. Jia, and A. Kuijper.
\newblock A light and faster regional convolutional neural network for object
  detection in optical remote sensing images.
\newblock {\em ISPRS J. P\&RS}, 141:208 -- 218, 2018.

\bibitem{Fan_2010_TNN}
J. {Fan}, W. {Xu}, Y. {Wu}, and Y. {Gong}.
\newblock Human tracking using convolutional neural networks.
\newblock {\em IEEE Transactions on Neural Networks}, 21(10):1610--1623, 2010.

\bibitem{feng-isprs2021}
Jie Feng, Dening Zeng, Xiuping Jia, Xiangrong Zhang, Jie Li, Yuping Liang, and
  Licheng Jiao.
\newblock Cross-frame keypoint-based and spatial motion information-guided
  networks for moving vehicle detection and tracking in satellite videos.
\newblock {\em ISPRS Journal of Photogrammetry and Remote Sensing},
  177:116--130, 2021.

\bibitem{Harris88}
C. {Harris} and M. {Stephens}.
\newblock A combined corner and edge detector.
\newblock In {\em In Proc. of Fourth Alvey Vision Conference}, pages 147--151,
  1988.

\bibitem{he-iccv2015}
Kaiming He, Xiangyu Zhang, Shaoqing Ren, and Jian Sun.
\newblock Delving deep into rectifiers: Surpassing human-level performance on
  imagenet classification.
\newblock In {\em 2015 IEEE International Conference on Computer Vision
  (ICCV)}, pages 1026--1034, 2015.

\bibitem{he-cvpr2016}
Kaiming He, Xiangyu Zhang, Shaoqing Ren, and Jian Sun.
\newblock Deep residual learning for image recognition.
\newblock In {\em 2016 IEEE Conference on Computer Vision and Pattern
  Recognition (CVPR)}, pages 770--778, 2016.

\bibitem{imbert-thesis2019}
J. Imbert.
\newblock Fine-tuning of fully convolutional networks for vehicle detection in
  satellite images: Data augmentation and hard examples mining.
\newblock Master's thesis, KTH, 2019.

\bibitem{karpathy2014large}
Andrej Karpathy, George Toderici, Sanketh Shetty, Thomas Leung, Rahul
  Sukthankar, and Li Fei-Fei.
\newblock Large-scale video classification with convolutional neural networks.
\newblock In {\em Proceedings of the IEEE conference on Computer Vision and
  Pattern Recognition}, pages 1725--1732, 2014.

\bibitem{koga-rs2018}
Y. Koga, H. Miyazaki, and R. Shibasaki.
\newblock A cnn-based method of vehicle detection from aerial images using hard
  example mining.
\newblock {\em Remote Sensing}, 10, Jan. 2018.

\bibitem{kopsiaftis-igarss2015}
G. {Kopsiaftis} and K. {Karantzalos}.
\newblock Vehicle detection and traffic density monitoring from very high
  resolution satellite video data.
\newblock In {\em IGARSS}, July 2015.

\bibitem{krizhevsky-2009}
Alex Krizhevsky.
\newblock Learning multiple layers of features from tiny images.
\newblock Master's thesis, Computer Science Department, 2009.

\bibitem{lalonde-cvpr2018}
R. LaLonde, D. Zhang, and M. Shah.
\newblock Clusternet: Detecting small objects in large scenes by exploiting
  spatio-temporal information.
\newblock In {\em CVPR}, June 2018.

\bibitem{li-JARS2019}
H. Li, L. Chen, F. Li, and M. Huang.
\newblock Ship detection and tracking method for satellite video based on
  multiscale saliency and surrounding contrast analysis.
\newblock {\em Journal of Applied Remote Sensing}, 13(2):1 -- 17, 2019.

\bibitem{liu-tip2015}
X. {Liu}, G. {Zhao}, J. {Yao}, and C. {Qi}.
\newblock Background subtraction based on low-rank and structured sparse
  decomposition.
\newblock {\em IEEE TIP}, 24(8):2502--2514, Aug 2015.

\bibitem{Long_2019_CVPR}
Fuchen Long, Ting Yao, Zhaofan Qiu, Xinmei Tian, Jiebo Luo, and Tao Mei.
\newblock Gaussian temporal awareness networks for action localization.
\newblock In {\em The IEEE Conference on Computer Vision and Pattern
  Recognition (CVPR)}, June 2019.

\bibitem{lowe1999object}
David~G Lowe.
\newblock Object recognition from local scale-invariant features.
\newblock In {\em Proceedings of the seventh IEEE international conference on
  computer vision}, volume~2, pages 1150--1157. Ieee, 1999.

\bibitem{Luo_2019_ICCV}
Chenxu Luo and Alan~L. Yuille.
\newblock Grouped spatial-temporal aggregation for efficient action
  recognition.
\newblock In {\em The IEEE International Conference on Computer Vision (ICCV)},
  October 2019.

\bibitem{meng}
L. {Meng} and J.~P. {Kerekes}.
\newblock Object tracking using high resolution satellite imagery.
\newblock {\em IEEE J. of Selected Topics in Applied Earth Observations and
  Remote Sensing}, 5(1):146--152, 2 2012.

\bibitem{mou-igarss2016}
L. {Mou} and X.~X. {Zhu}.
\newblock Spatiotemporal scene interpretation of space videos via deep neural
  network and tracklet analysis.
\newblock In {\em IGARSS}, July 2016.

\bibitem{muja2009fast}
Marius Muja and David~G Lowe.
\newblock Fast approximate nearest neighbors with automatic algorithm
  configuration.
\newblock {\em VISAPP (1)}, 2(331-340):2, 2009.

\bibitem{mundhenk-eccv2016}
T. Mundhenk, G. Konjevod, W. Sakla, and K. Boakye.
\newblock A large contextual dataset for classification, detection and counting
  of cars with deep learning.
\newblock In {\em ECCV}, Oct. 2016.

\bibitem{murthy2014skysat}
Kiran Murthy, Michael Shearn, Byron~D Smiley, Alexandra~H Chau, Josh Levine,
  and M~Dirk Robinson.
\newblock Skysat-1: very high-resolution imagery from a small satellite.
\newblock In {\em Sensors, Systems, and Next-Generation Satellites XVIII},
  volume 9241, page 92411E. International Society for Optics and Photonics,
  2014.

\bibitem{otsu-tsmc1979}
Nobuyuki Otsu.
\newblock A threshold selection method from gray-level histograms.
\newblock {\em IEEE Transactions on Systems, Man, and Cybernetics},
  9(1):62--66, 1979.

\bibitem{paszke-neurips2019}
Adam Paszke, Sam Gross, Francisco Massa, Adam Lerer, James Bradbury, Gregory
  Chanan, Trevor Killeen, Zeming Lin, Natalia Gimelshein, Luca Antiga, Alban
  Desmaison, Andreas K\"{o}pf, Edward Yang, Zach DeVito, Martin Raison, Alykhan
  Tejani, Sasank Chilamkurthy, Benoit Steiner, Lu Fang, Junjie Bai, and Soumith
  Chintala.
\newblock {\em PyTorch: An Imperative Style, High-Performance Deep Learning
  Library}.
\newblock Curran Associates Inc., Red Hook, NY, USA, 2019.

\bibitem{perez2017effectiveness}
Luis Perez and Jason Wang.
\newblock The effectiveness of data augmentation in image classification using
  deep learning.
\newblock {\em arXiv preprint arXiv:1712.04621}, 2017.

\bibitem{pflugfelder2020learning}
Roman Pflugfelder, Axel Weissenfeld, and Julian Wagner.
\newblock On learning vehicle detection in satellite video.
\newblock {\em arXiv preprint arXiv:2001.10900}, 2020.

\bibitem{redmon-corr2018}
J. Redmon and A. Farhadi.
\newblock Yolov3: An incremental improvement.
\newblock {\em CoRR}, abs/1804.02767, 2018.

\bibitem{reilly2010detection}
Vladimir Reilly, Haroon Idrees, and Mubarak Shah.
\newblock Detection and tracking of large number of targets in wide area
  surveillance.
\newblock In {\em European conference on computer vision}, pages 186--199.
  Springer, 2010.

\bibitem{ren-nips2015}
S. Ren, K. He, R. Girshick, and J. Sun.
\newblock Faster r-cnn: Towards real-time object detection with region proposal
  networks.
\newblock In {\em NIPS}, 2015.

\bibitem{simonyan2014very}
Karen Simonyan and Andrew Zisserman.
\newblock Very deep convolutional networks for large-scale image recognition.
\newblock {\em arXiv preprint arXiv:1409.1556}, 2014.

\bibitem{sommer2016survey}
Lars~Wilko Sommer, Michael Teutsch, Tobias Schuchert, and J{\"u}rgen Beyerer.
\newblock A survey on moving object detection for wide area motion imagery.
\newblock In {\em 2016 IEEE Winter Conference on Applications of Computer
  Vision (WACV)}, pages 1--9. IEEE, 2016.

\bibitem{springenberg-iclrw2015}
Jost~Tobias Springenberg, Alexey Dosovitskiy, Thomas Brox, and Martin~A.
  Riedmiller.
\newblock Striving for simplicity: The all convolutional net.
\newblock In Yoshua Bengio and Yann LeCun, editors, {\em 3rd International
  Conference on Learning Representations, {ICLR} 2015, San Diego, CA, USA, May
  7-9, 2015, Workshop Track Proceedings}, 2015.

\bibitem{Szegedy_2015_CVPR}
Christian Szegedy, Wei Liu, Yangqing Jia, Pierre Sermanet, Scott Reed, Dragomir
  Anguelov, Dumitru Erhan, Vincent Vanhoucke, and Andrew Rabinovich.
\newblock Going deeper with convolutions.
\newblock In {\em Proceedings of the IEEE Conference on Computer Vision and
  Pattern Recognition (CVPR)}, June 2015.

\bibitem{tran2015learning}
Du Tran, Lubomir Bourdev, Rob Fergus, Lorenzo Torresani, and Manohar Paluri.
\newblock Learning spatiotemporal features with 3d convolutional networks.
\newblock In {\em Proceedings of the IEEE international conference on computer
  vision}, pages 4489--4497, 2015.

\bibitem{wagner-phd2020}
Julian Wagner.
\newblock {\em Detecting moving vehicles in satellite videos using deep neural
  networks}.
\newblock PhD thesis, University of Technology Vienna, 2020.

\bibitem{guo-rs2018}
G. Wei, Y. Wen, Z. Haijian, and H. Guang.
\newblock Geospatial object detection in high resolution satellite images based
  on multi-scale convolutional neural network.
\newblock {\em Remote Sensing}, 10(1), 2018.

\bibitem{xu-jrsgis2017}
A. Xu, J. Wu, G. Zhang, S. Pan, T. Wang, Y. Jang, and X. Shen.
\newblock Motion detection in satellite video.
\newblock {\em Journal of Remote Sensing and GIS}, 6(2):1--9, 2017.

\bibitem{yang-iccv2019}
F. Yang, H. Fan, P. Chu, E. Blasch, and H. Ling.
\newblock Clustered object detection in aerial images.
\newblock In {\em ICCV}, 2019.

\bibitem{yang-corr2018}
M.~Y. Yang, W. Liao, X. Li, and B. Rosenhahn.
\newblock Vehicle detection in aerial images.
\newblock {\em CoRR}, abs/1801.07339, 2018.

\bibitem{yang-sensors2016}
T. Yang, X. Wang, B. Yao, J. Li, Y. Zhang, Z. He, and W. Duan.
\newblock Small moving vehicle detection in a satellite video of an urban area.
\newblock {\em Sensors}, 16(9):1528, Sept. 2016.

\bibitem{zhang-corr2019a}
J. Zhang, X. Jia, and J. Hu.
\newblock Error bounded foreground and background modeling for moving object
  detection in satellite videos.
\newblock {\em CoRR}, 2019.

\bibitem{zhang-GSRS2019}
J. {Zhang}, X. {Jia}, and J. {Hu}.
\newblock Error bounded foreground and background modeling for moving object
  detection in satellite videos.
\newblock {\em IEEE Transactions on Geoscience and Remote Sensing}, pages
  1--11, 2019.

\bibitem{zhang-rsens2019}
J. Zhang, X. Jia, and J. Hu.
\newblock Local region proposing for frame-based vehicle detection in satellite
  videos.
\newblock {\em Remote Sensing}, 11(20):2372, Oct. 2019.

\bibitem{zhang-corr2019b}
Junpeng Zhang, Xiuping Jia, Jiankun Hu, and Jocelyn Chanussot.
\newblock Online structured sparsity-based moving object detection from
  satellite videos.
\newblock {\em CoRR}, 2019.

\bibitem{zhang-tgars2020}
Junpeng Zhang, Xiuping Jia, Jiankun Hu, and Jocelyn Chanussot.
\newblock Online structured sparsity-based moving-object detection from
  satellite videos.
\newblock {\em IEEE Transactions on Geoscience and Remote Sensing},
  58(9):6420--6433, 2020.

\bibitem{zhang-tpami2021}
Junpeng Zhang, Xiuping Jia, Jiankun Hu, and Kun Tan.
\newblock Moving vehicle detection for remote sensing video surveillance with
  nonstationary satellite platform.
\newblock {\em IEEE Transactions on Pattern Analysis and Machine Intelligence},
  pages 1--1, 2021.

\bibitem{zhang-spie2017}
X. Zhang and J. Xiang.
\newblock Moving object detection in video satellite image based on deep
  learning.
\newblock In {\em LIDAR Imaging Detection and Target Recognition}, volume
  10605, pages 1149 -- 1156. SPIE, 2017.

\bibitem{zhao-tgars2022}
Manqi Zhao, Shengyang Li, Shiyu Xuan, Longxuan Kou, Shuai Gong, and Zhuang
  Zhou.
\newblock Satsot: A benchmark dataset for satellite video single object
  tracking.
\newblock {\em IEEE Transactions on Geoscience and Remote Sensing}, pages 1--1,
  2022.

\bibitem{zhou-icml2011}
T. Zhou and D. Tao.
\newblock Godec: Randomized low-rank \& sparse matrix decomposition in noisy
  case.
\newblock In {\em ICML}, 2011.

\bibitem{zhou-tpami2013}
X. {Zhou}, C. {Yang}, and W. {Yu}.
\newblock Moving object detection by detecting contiguous outliers in the
  low-rank representation.
\newblock {\em IEEE TPAMI}, 35(3):597--610, March 2013.

\end{thebibliography}
}

%

\begin{IEEEbiography}[{\includegraphics[width=1in,height=1.25in,clip,keepaspectratio]{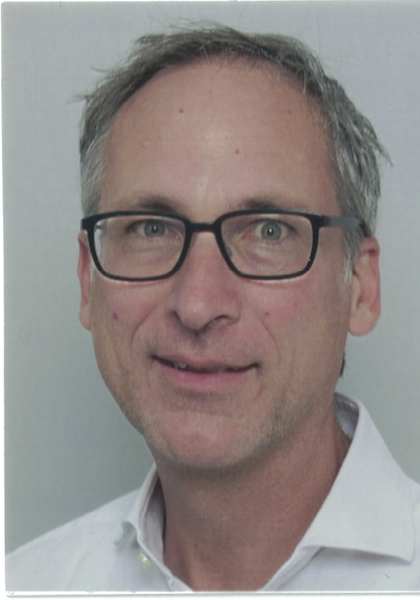}}]{Roman Pflugfelder}
is Fellow of the Technical University of Munich and the Technion, Israel, Scientist at the AIT Austrian Institute of Technology and lecturer at TU Wien. He received in 2002 a MSc degree in Informatics at TU Wien and in 2008 a PhD in Telematics at the TU Graz, Austria. In 2001, he received the Kurt Gödel stipend from TU Wien for an academic visit to the Queensland University of Technology, Australia. His research aims at visual motion analysis, tracking, recognition, and learning applied to automated video surveillance.
Roman contributed with 69 papers and patents to research fields such as camera calibration, object detection, object tracking and event recognition. Roman co-organised the Visual Object Tracking Challenges VOT'13-14 and VOT'16-21 and was program chair of AVSS'15. Currently he is steering committee member of AVSS. He is regular reviewer for major computer vision / machine learning conferences and journals. He received in 2008 a WWTF Career Grant, in 2014 the WACV Best Paper Award, in 2019 the CVPR Outstanding Reviewer Award and 2021 a Marie Curie Fellowship.
\end{IEEEbiography}
\begin{IEEEbiography}[{\includegraphics[width=1in,height=1.25in,clip,keepaspectratio]{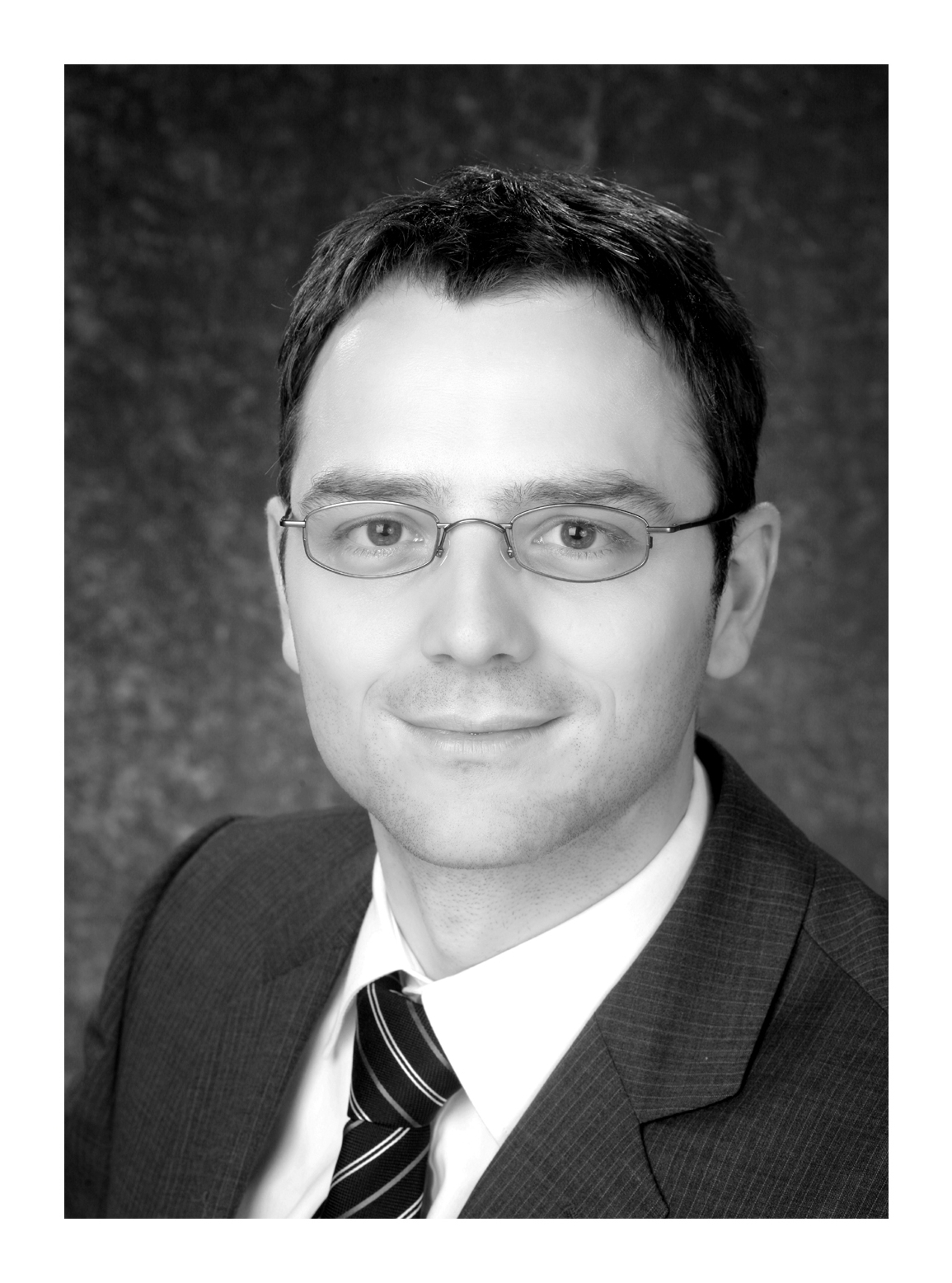}}]{Axel Weissenfeld}
is a member of the competence unit Sensing and Vision Solutions at AIT. He is currently working on vehicle detection in satellite video data and contactless fingerprint detection for biometric applications. His research interests lie at the intersection of computer vision, machine learning, and human-computer interaction. Before Axel joined AIT in 2012, he was working on several real-time hardware platforms with the focus on intelligent video and forensic analysis for Bosch Security Systems. From 2003 to 2007, he worked as a research assistant at the Institute of Information Processing, Leibniz University Hannover. His work was directed towards computer vision, human-machine interfaces, multiple view geometry and computer graphics. He received his Dipl.-Ing. and Dr.-Ing. degree from the Leibniz University Hannover in 2003 and 2010, respectively.
\end{IEEEbiography}
\begin{IEEEbiography}[{\includegraphics[width=1in,height=1.25in,clip,keepaspectratio]{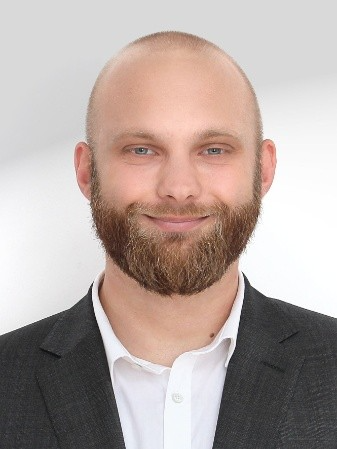}}]{Julian Wagner}
is a self-employed Computer Vision Engineer. He received 2017 a BSc degree and in 2020 a MSc degree in Informatics at TU Wien. His interest is on Machine Learning in Visual Computing which led to his master's thesis with the title {\em Detecting moving vehicles in satellite videos using deep neural networks}. Julian also co-authored in 2020 a paper at the $25^{th}$ Computer Vision Winter Workshop with the title {\em On Learning Vehicle Detection in Satellite Video}.
\end{IEEEbiography}




\end{document}